\documentclass{article} 

\PassOptionsToPackage{hyphens}{url}

\usepackage{iclr2023_conference,times}

\usepackage{amsmath,amsfonts,bm}

\def\eqref#1{equation~\ref{#1}}

\def\1{\bm{1}}

\DeclareMathAlphabet{\mathsfit}{\encodingdefault}{\sfdefault}{m}{sl}
\SetMathAlphabet{\mathsfit}{bold}{\encodingdefault}{\sfdefault}{bx}{n}

\usepackage{amssymb}
\usepackage{microtype}
\usepackage{amsmath}
\usepackage{booktabs}
\usepackage{colortbl}
\usepackage[utf8]{inputenc}
\definecolor{lightgray}{rgb}{0.9,0.9,0.9}
\usepackage{caption}
\usepackage{subcaption}
\usepackage{xcolor}
\usepackage{graphicx}
\usepackage{setspace}
\usepackage[hidelinks]{hyperref}
\usepackage{url}
\usepackage{multirow}
\usepackage{colortbl}
\usepackage{tabularx}
\usepackage{blindtext}
\usepackage{pgfplots}
\pgfplotsset{compat=1.18} 
\usepackage{tikz}
\usetikzlibrary{er,positioning,bayesnet}
\usepackage[inline]{enumitem}
\usepackage{makecell}
\usepackage{siunitx}
\usepackage{nicefrac}
\usepackage{listings}
\usepackage[raster,skins]{tcolorbox} 
\usepackage{adjustbox}
\usepackage{diagbox}
\usepackage{color}
\usepackage{array}
\usepackage{longtable}
\usepackage{amsmath}
\usepackage{booktabs}
\usepackage{makecell}
\usepackage{multirow}

\title{Qwen2.5-Math Technical Report: \\ Toward Mathematical Expert Model via Self-Improvement}
\author{
\\
\parbox{\linewidth}{An Yang, Beichen Zhang, Binyuan Hui, Bofei Gao, Bowen Yu$^{\dag}$, Chengpeng Li, Dayiheng Liu$^{\dag}$, Jianhong Tu, Jingren Zhou, Junyang Lin$^{\dag}$, Keming Lu, Mingfeng Xue, Runji Lin, Tianyu Liu, Xingzhang Ren, Zhenru Zhang}
\AND
Qwen Team, Alibaba Group\thanks{Authors are ordered alphabetically by the first name. $^\dag$Corresponding author.} \\
}

\clearpage

\iclrfinalcopy
\begin{document}

\maketitle

\begin{abstract}

In this report, we present a series of math-specific large language models: Qwen2.5-Math and Qwen2.5-Math-Instruct-1.5B/7B/72B.
The core innovation of the Qwen2.5 series lies in integrating the philosophy of self-improvement throughout the entire pipeline, from pre-training and post-training to inference:
(1) During the pre-training phase, Qwen2-Math-Instruct is utilized to generate large-scale, high-quality mathematical data.
(2) In the post-training phase, we develop a reward model (RM) by conducting massive sampling from Qwen2-Math-Instruct. This RM is then applied to the iterative evolution of data in supervised fine-tuning (SFT). With a stronger SFT model, it's possible to iteratively train and update the RM, which in turn guides the next round of SFT data iteration. 
On the final SFT model, we employ the ultimate RM for reinforcement learning, resulting in the Qwen2.5-Math-Instruct.
(3) Furthermore, during the inference stage, the RM is used to guide sampling, optimizing the model's performance.


Qwen2.5-Math-Instruct supports both Chinese and English, and possess advanced mathematical reasoning capabilities, including Chain-of-Thought (CoT) and Tool-Integrated Reasoning (TIR).
We evaluate our models on 10 mathematics datasets in both English and Chinese, such as GSM8K, MATH, GaoKao, AMC23, and AIME24, covering a range of difficulties from grade school level to math competition problems. 
The flagship model, Qwen2.5-Math-72B-Instruct, significantly outperforms both open-source models and leading closed-source models (e.g., GPT-4o, Gemini Math-Specialized 1.5 Pro).
Particularly in the challenging AMC 2023, with the assistance of RM, Qwen2.5-Math-72B-Instruct successfully solves almost all the problems.
Qwen2.5-Math-7B-Instruct surpasses Qwen2-Math-Instruct 72B in performance.
Under CoT and TIR settings, it achieves MATH scores of 83.6 and 85.3, respectively. 
Even our smallest 1.5B model, achieving a MATH score of around 80 when utilizing the Python Interpreter, outperforms the majority of current models in this domain.
We hope that Qwen2.5-Math can contribute to the community for solving complex mathematical problems. 

The base models, instruct models, and reward model of the Qwen2.5-Math series are available on Hugging Face~\footnote{\url{https://huggingface.co/Qwen}} and ModelScope\footnote{\url{https://modelscope.cn/organization/qwen}}, and the evaluation scripts on GitHub\footnote{\url{https://github.com/QwenLM/Qwen2-Math}}. 
We have also developed a demo that supports the TIR mode in Qwen-Agent\footnote{\url{https://github.com/QwenLM/Qwen-Agent}}, which allows running code locally to experience Tool-Integrated Reasoning capabilities of Qwen2.5-Math.

\end{abstract}

\clearpage

\tableofcontents
\clearpage

\section{Introduction}

\begin{figure}[htbp]
    \centering
    \includegraphics[width=1.0\columnwidth]{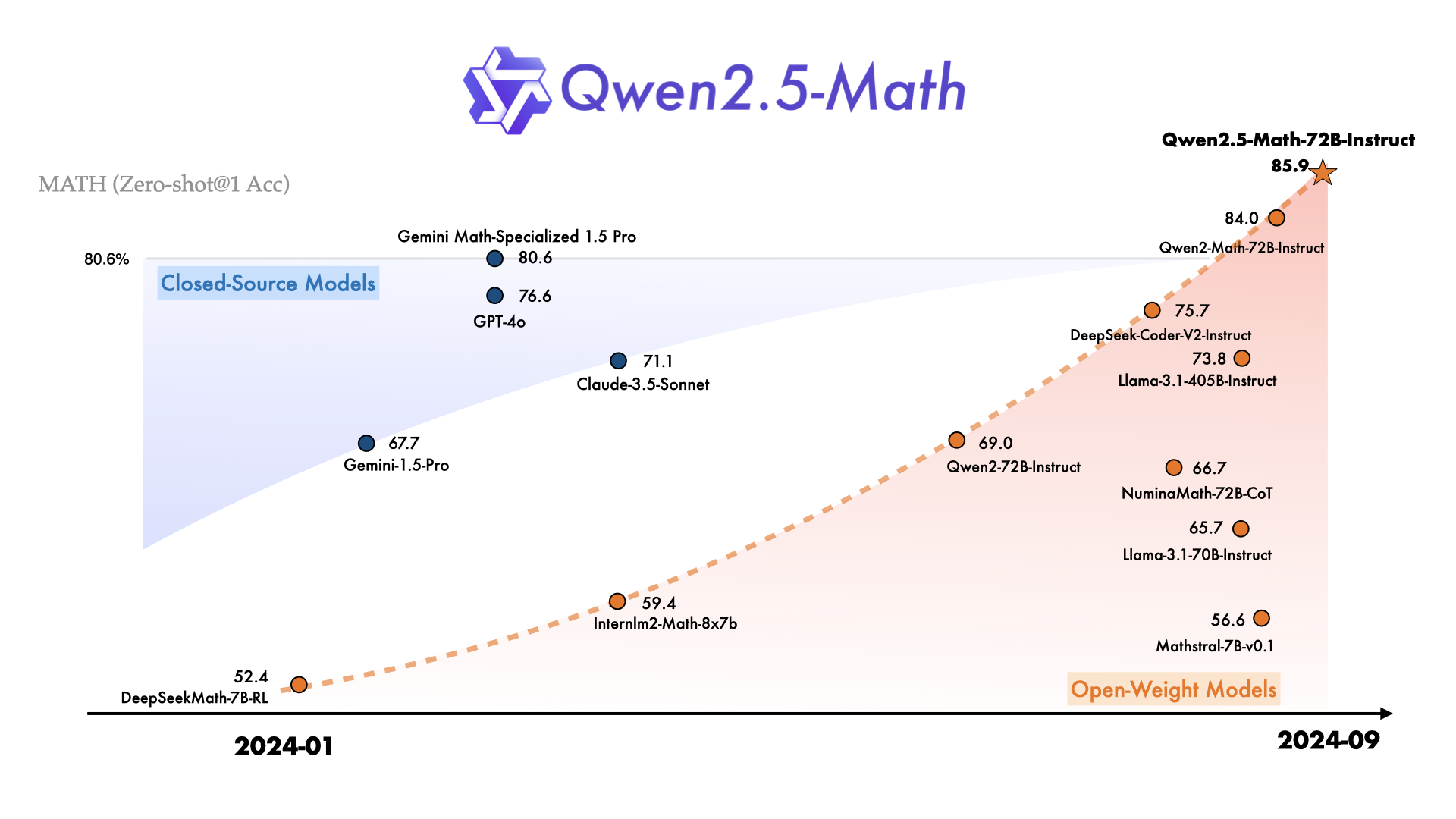}
    \vspace{-1mm}
    \caption{The pass@1 performance of Qwen2.5-Math-72B-Instruct on MATH by the Chain-of-Thought reasoning.}
    \label{fig:intro}
\end{figure}

\label{sec:intro}
Over the past year, we have devoted considerable effort to researching and enhancing the reasoning capabilities of large language models, with a particular emphasis on their ability to solve arithmetic and mathematical problems. 
In this report, we introduce a series of math-specific large language models, Qwen2.5-Math, Qwen2.5-Math-RM, and Qwen2.5-Math-Instruct-1.5B/7B/72B. 
To provide a comprehensive understanding of the technical developments behind Qwen2.5-Math, we also offer a detailed overview of its predecessor, Qwen2-Math~\citep{qwen2math}.

We introduce a series of self-improvement techniques to develop Qwen2.5-Math models on top of the Qwen2-Math. Self-improvement techniques take advantage of supervision from large language models themselves~\citep{cao2024towards}. Specifically, we apply self-improvement from three aspects during the training of Qwen2.5-Math. In pre-training, we employ Qwen2-Math-Instruct to synthesize math queries and corresponding responses on a large scale to enrich the pre-training corpus of Qwen2.5-Math.
In post-training, we train a reward model on massive sampling from previous models and apply it to the iterative evolution of data in supervised fine-tuning.
The better mathematical models trained from this enhancement lead to a more robust reward model, Qwen2.5-Math-RM.
Then, we use this reward model in reinforcement learning and best-of-N sampling during inference.
Synthetic data and judgment play a significant role in the enhancement of Qwen2.5-Math compared with its predecessor.

\begin{figure}[htbp]
    \centering
\includegraphics[width=1\columnwidth]{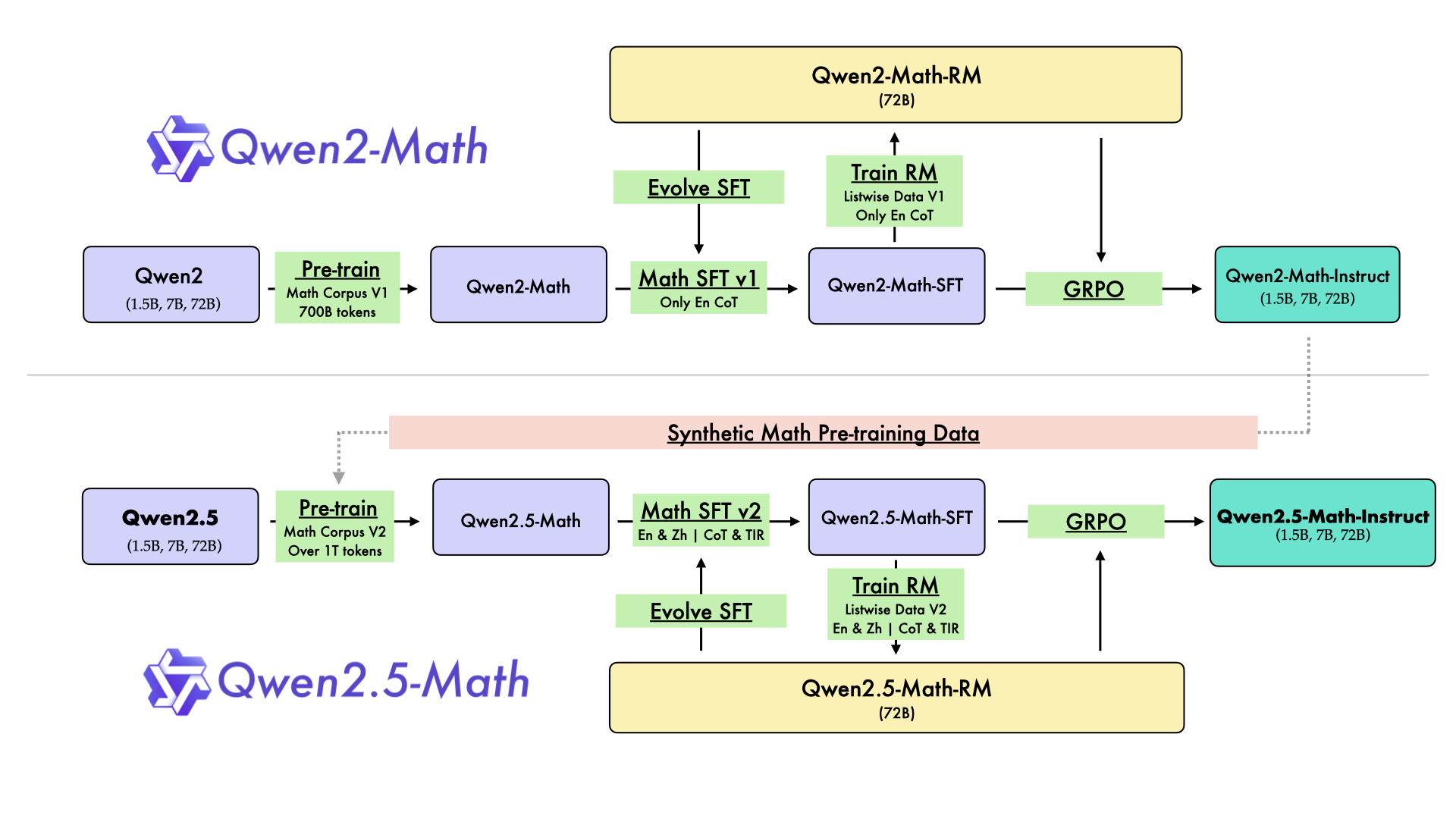}
    \vspace{-1mm}
    \caption{The development pipelines of Qwen2-Math and Qwen2.5-Math. }
    \label{fig:pipeline}
\end{figure}

Specifically, the overall pipelines for developing Qwen2-Math and Qwen2.5-Math are illustrated in Figure~\ref{fig:pipeline}.
First, the Qwen2-Math base models are trained on a high-quality mathematical pre-training dataset called the \textit{Qwen Math Corpus v1}, which contains approximately 700 billion tokens.
Second, we train a math-specific reward model Qwen2-Math-RM, derived from Qwen2-Math-72B, to create the Qwen2-Math-Instruct models. This reward model is used to construct Supervised Fine-Tuning (SFT) data through Rejection Sampling~\citep{rft}.
Moreover, the reward model plays a key role in the reinforcement learning stage, where we employ Group Relative Policy Optimization (GRPO)~\citep{deepseekmath} following SFT.
Third, leveraging the Qwen2-Math-72B-Instruct model, we synthesize additional high-quality mathematical pre-training data, which serves as the foundation for \textit{Qwen Math Corpus v2}. 
This updated corpus contains over 1 trillion tokens and is used to pre-train the Qwen2.5-Math models.
Lastly, similar to the process used for the Qwen2-Math-Instruct models, we construct the Qwen2.5-Math-RM and Qwen2.5-Math-Instruct models. 
An important distinction in this stage is the inclusion of both English and Chinese Chain-of-Thought (CoT) reasoning data, as well as Tool-Integrated Reasoning (TIR) data, for training the Qwen2.5-Math-Instruct models, as opposed to using only English CoT data as was done for Qwen2-Math-Instruct.

We evaluate our math-specific models on eight English and Chinese math benchmarks. 
Notably, the Qwen2.5-Math-7B base model achieves scores of 91.6, 55.4, and 57.6 on GSM8K~\citep{gsm8k}, MATH~\citep{math}, and GaoKao Math Cloze~\citep{gaokao-bench}, respectively, outperforming the Qwen2-72B~\citep{qwen2} general model, which achieves scores of 89.5, 51.1, and 55.9 on the same datasets. 
Additionally, the Qwen2.5-Math-72B base model sets a new state-of-the-art on the MATH benchmark, achieving a score of 66.8—an improvement of 5.3 points over Qwen2-Math-72B and 15.7 points over Qwen2-72B.

For the Instruct models, in CoT mode, the Qwen2.5-Math-1.5B-Instruct model surpasses the performance of all currently available open-source models on most metrics, including models as large as 70B parameters. 
Furthermore, the Qwen2.5-Math-7B-Instruct model nearly matches the performance of the Qwen2-Math-72B-Instruct model, indicating that improvements to the training data and strategy can, to a certain extent, compensate for the scaling up of parameters.
The Qwen2.5-Math-72B-Instruct model outperforms the Qwen2-Math-72B-Instruct model by an average margin of 4.4 and 6.1 points in English and Chinese, respectively, establishing itself as the best open-source mathematical model currently available. 
Moreover, all model sizes demonstrate significant improvements in their Chinese math problem-solving capabilities.
In our newly introduced TIR mode, performance sees further enhancement compared to CoT. 
For instance, the 72B model achieves close to 90 points on the MATH benchmark, and even the 1.5B model scores around 80, demonstrating that Qwen2.5 is now highly proficient at leveraging the Python Interpreter for accurate mathematical computation.

\section{Qwen2.5-Math Pre-training}

\label{sec:pre}

In mathematical pre-training, our primary focus is on constructing a high-quality dataset rich in mathematical content. 
This dataset encompasses a wide variety of sources, including math-related web texts, code snippets, encyclopedias, exam questions, and synthetic mathematical data generated by Qwen2~\citep{qwen2}.
The process of assembling this pre-training dataset involves several key steps: data recall, deduplication, filtering, data synthesis, and optimization of the data mixture. The final curated dataset, which forms the foundation of our pre-training, is termed the \textit{Qwen Math Corpus v1}.
The Qwen2-Math base models, initialized with Qwen2-1.5B/7B/72B, undergo continuous pre-training using the \textit{Qwen Math Corpus v1}.

\par

Prior to the construction of \textit{Qwen Math Corpus v1}, we observe that the suboptimal performance of general language models in mathematical reasoning stems from an insufficiency of mathematical data during pre-training. The existing endeavors pre-training to large-scale, specialized LLMs focused on mathematics~\citep{deepseekmath,internlmmath,lewkowycz2022solving,llemma} have unequivocally demonstrated the value of extracting a considerable corpus of mathematical texts from digital databases. Our initial strategy involves the recall of mathematical data from web sources, such as Common Crawl, to escalate the quantity of data. Concretely, we train a FastText~\citep{fasttext} classifier utilizing high-quality mathematical seed data and general text data. We leverage iterative training with more math data each epoch to continuously enhance the performance of the classifier. To recognize the missing mathematical-related data in the corpus pool, we leverage meta-information, such as URLs, from the recalled data to expand the data pool for mathematical data retrieval. Subsequently, deduplication techniques, including MinHash~\citep{minhash}, are employed to filter out similar mathematical documents.
\par

Upon collecting a substantial volume of mathematical data, our focus shifts toward enhancing its quality. 
For this, we implement a language-model-based filtering technique to further curate the dataset.
Specifically, we utilize the Qwen2-0.5B-Instruct model~\citep{qwen2}, augmented with prompt engineering, to evaluate the quality of potential data entries. 
Data that receive higher scores, indicating higher quality according to the language model, are prioritized for inclusion in the final dataset.
Beyond recalling a diverse set of mathematical documents and filtering out low-quality data, we draw inspiration from previous efforts in generating synthetic mathematical data~\citep{mammoth2,jiuzhang3}. 
We employ the Qwen2-72B-Instruct model to synthesize a large amount of mathematical pre-training corpus. 
At this stage, the high-quality mathematical data already collected are used as reference materials. Using the Qwen2-72B-Instruct model, we: (1) extract and refine existing mathematical question-answer data from these references, and (2) directly generate new mathematical question-answer pairs.
\par

In the final phase, we conduct ablation studies on data mixture using a small math-specific language model, Qwen2-Math-1.5B.
Based on the findings, we construct the \textit{Qwen Math Corpus v1}, which comprises 700 billion tokens in total. 
We initialize the Qwen2-Math-1.5B/7B/72B pre-training with intermediate checkpoints from the corresponding Qwen2-1.5B/7B/72B base models. 
These models are then continuously pre-trained on \textit{Qwen Math Corpus v1} with a context length of 4K.

\par

Following the training of the Qwen2-Math base models, we further upgrade them to Qwen2.5-Math models through three primary avenues:
(1) We utilize the Qwen2-Math-72B-Instruct model, further post-trained with the steps described in Section~\ref{sec:post}, to synthesize additional high-quality mathematical pre-training data. 
2) We aggregate more high-quality mathematical data, especially in Chinese, sourced from web documents, books, and code repositories across multiple recall cycles. 
As a result of these efforts, we compile the \textit{Qwen Math Corpus v2} for Qwen2.5-Math-1.5B/7B/72B pre-training, while maintaining a context length of 4K. 
Compared to \textit{Qwen Math Corpus v1}, the total token count of \textit{Qwen Math Corpus v2} escalates from 700B to over 1T.
(3) Instead of initializing from the Qwen2 series, we leverage the Qwen2.5 series base models for parameter initialization, as they exhibit enhanced capabilities in language understanding, code generation, and text reasoning. 
Qwen2.5-Math models are continuously pre-trained on \textit{Qwen Math Corpus v2} under a math pre-training setup similar to Qwen2-Math.
Benefiting from the improvements in both the dataset and the base model, Qwen2.5-Math models demonstrate further advancements in mathematical reasoning abilities beyond Qwen2-Math.
\section{Qwen2.5-Math Post-training}

\label{sec:post}

After completing extensive mathematical pre-training, we proceed with post-training to further augment the mathematical logical reasoning capabilities of Qwen-Math, specifically focusing on Chain-of-Thought (CoT) and Tool-Integrated Reasoning (TIR).
Our investigation is particularly focused on two key challenges:
(1) How to automatically generate a substantial volume of high-quality and reliable CoT and TIR annotations, and
(2) How to effectively leverage these annotations for both Supervised Fine-Tuning and Reinforcement Learning.

\subsection{Supervised Fine-tuning}

We aim for Qwen-Math to excel in two core capabilities: solving math problems through step-by-step natural language reasoning~\citep{cot}, and leveraging external tools (e.g., a Python interpreter) to address complex mathematical or algorithmic reasoning tasks~\citep{yue2023mammoth}.
We have constructed dedicated datasets for both Chain-of-Though~(CoT) and Tool-integrated Reasoning~(TIR) and combined these datasets to train the model jointly.
All models are trained for 3 epochs with a sequence length of 4,096 tokens. 
For the 72B model, we use a batch size of 256 and a learning rate of $5\times 10^{-6}$.
For the 1.5B and 7B models, we set the batch size to 128 and the learning rate to $2\times 10^{-5}$.
During training, the learning rate gradually decays to a final value of $7\times 10^{-7}$.

\subsubsection{Chain-of-Thought Data Synthesis}

\paragraph{Query Construction.}
The chain-of-thought dataset comprises a wide-ranging collection of 580K English and 500K Chinese mathematical problems, including both annotated and synthesized items. 
The annotated problems are derived from well-established sources such as the training set of GSM8K~\citep{gsm8k}, MATH~\citep{math}, and NuminaMath~\citep{numinamath}.
In an effort to bolster the Chinese reasoning capabilities of Qwen2.5-Math, we have further enriched the dataset with additional Chinese mathematical problems from exclusive K-12 problem collections.
The synthesized problems are evolved from the annotated ones using the MuggleMath approach~\citep{mugglemath}.
To maintain a balanced distribution across varying levels of problem complexity, we utilize a difficulty-scoring model to categorize our problem set effectively.

\paragraph{Response Construction.}

We adopt an iterative approach that leverages rejection sampling, guided by reward modeling and annotated answers, to incrementally enhance the quality of responses~\citep{rft}.
At each iteration, the current best model is deployed to generate multiple reasoning pathways for the given problems, expanding the pool of candidate solutions.
For problems with annotated answers, we select the top-$k$ reasoning paths with correct final answers from the pool.
For synthesized problems lacking definitive answers, we implement a weighted majority voting mechanism to deduce the most plausible correct reasoning paths.
From these, we choose the top-$k$ pathways that receive the highest reward scores.
In the development of Qwen2.5-Math, an additional iteration is conducted using the Qwen2-Math-Instruct models to polish the quality of responses further. 
The final CoT training set encompasses 2000K English samples and 500K Chinese samples.

\subsubsection{Tool-integrated Reasoning Data Synthesis}

It is important to recognize that while CoT prompting plays a crucial role in enhancing the reasoning skills of large language models, it faces challenges in achieving computational accuracy and in handling complex mathematical or algorithmic problems, such as finding the roots of quadratic equations or computing the eigenvalues of matrices~\citep{yue2023mammoth}. 
To overcome these limitations and improve the model’s proficiency in precise calculations, symbolic manipulation, and algorithmic reasoning, we have developed a dataset that incorporates a tool-integrated reasoning format. This innovative format enables the model to leverage a Python interpreter as an auxiliary resource in reasoning tasks.

\paragraph{Query Construction.} 
The tool-integrated reasoning dataset consists of 190K annotated problems and 205K synthesized problems. 
The annotated problems are sourced from the training sets of established benchmarks, including GSM8K~\citep{gsm8k}, MATH~\citep{math}, CollegeMath~\citep{collegeMATH}, and NuminaMath~\citep{numinamath}.
The synthesized problems are generated by employing techniques from MuggleMath~\citep{mugglemath} and DotaMath~\citep{dotamath} designed to facilitate query evolution within the GSM8K and MATH training sets. 
Additionally, we have selected 75K annotated problems for translation into Chinese using the Qwen2-72B model~\citep{qwen2}, aimed at enhancing the model’s reasoning capabilities in Chinese.

\paragraph{Response Construction.}

For the annotated problems, we utilize an online Rejection Fine-Tuning (RFT)~\citep{rft,onlinerft} approach to iteratively generate tool-integrated reasoning paths whose final answers align with the reference answers. 
In each RFT iteration, we carry out multiple nucleus samplings with the currently best model at various temperatures, increasing the sample size for particularly challenging problems. 
After each iteration, to enhance data diversity, we apply a deduplication process to the responses, and the resulting cleaned dataset is then used to fine-tune the model for the next iteration.
For the synthesized problems, we employ the optimal model derived from the online RFT process to generate reasoning samples.
Majority voting is employed to select the most probable correct reasoning paths, which are subsequently incorporated into the overall dataset.

\subsection{Reward Model Training}
To provide supervisory signals beyond merely the final answer during both the selection of supervised fine-tuning data and the subsequent stages of reinforcement learning training, we have developed a mathematical reward model for Qwen2-Math and Qwen2.5-Math, referred to as Qwen2-Math-RM and Qwen2.5-Math-RM, respectively. 
These reward models are specifically designed to guide the model throughout the training process by offering more granular feedback on the quality of reasoning and intermediate steps, ultimately facilitating more robust model improvements.

\subsubsection{Data Synthesis}

In the development of Qwen2-Math-RM, we utilize 206K English mathematical problems, each paired with 6 candidate responses sampled from an intermediate version of Qwen2-Math. 
For Qwen2.5-Math-RM, we further enhance its support for both the Chinese language and TIR mode, training it with a more diverse set of 361K English and 257K Chinese mathematical problems, with each problem accompanied by 6 responses sampled from Qwen2.5-Math. 
This expansion ensures that Qwen2.5-Math-RM is well-equipped to provide supervisory feedback across a broader range of problem types and languages.

To establish the preference signals among the responses, we check the final answers of the responses to determine their correctness. 
Responses with the correct answers are labeled as positive, while those with incorrect answers are labeled as negative, thereby naturally creating a ranking relationship among the responses. 
We then filter out any cases where all responses are either entirely correct or entirely incorrect. 
However, to avoid the potential drawback of retaining only overly simplistic data, we enrich the dataset with responses from various intermediate versions and models of different sizes. 
This strategy ensures a more balanced distribution of query difficulty and maintains an even ratio of positive to negative responses.

\subsubsection{Training Strategy}
We initialize the reward model from the supervised fine-tuning model. In terms of architecture, we replace the language modeling head originally used for next-token prediction with a scalar-value head, consisting of two linear layers.
As previously mentioned, each query in the reward model's training dataset is paired with $6$ responses, comprising both positive and negative candidates. 
If there are $k$ positive responses, then the remaining $6-k$ are negative. Following \cite{NEURIPS2022_b1efde53}, the loss function for the reward model can therefore be formulated as follows:

\begin{equation}\label{eq:reward_loss}
\mathcal{L}_{rm}(\theta) = -\frac{1}{k \times (6-k)} E_{(x,y_{pos},y_{neg}) \sim D} \left[ \log \left( \sigma \left( r_{\theta}(x,y_{pos}) - r_{\theta}(x,y_{neg}) \right) \right) \right].
\end{equation}

Here, $r_{\theta}(x,y)$ denotes the output of the reward model, where $x$ represents the problem and $y$ is the corresponding response. 
Rather than breaking these into multiple individual pairs and computing the loss in a pairwise fashion, we adopt a listwise approach to compute the ranking loss directly over valid pairs. This method enhances both training efficiency and effectiveness.

\subsection{Reinforcement Learning}

\paragraph{Query Selection.}

The queries for reinforcement learning training are selected from the reward model’s training set. 
We leverage supervised fine-tuning models with varying sizes to resample $8$ responses for each query, with each response classified as either correct or incorrect by comparing it to the gold-standard answer.
In the reinforcement learning stage, our primary goal is to ensure that the model consistently produces correct answers for queries where a correct response is possible.
Therefore, we only retain queries for which 2 to 5 out of the 8 responses are correct. Queries with fewer than 2 correct answers are excluded as they indicate that the current Math model lacks the fundamental capability to learn from them. 
Likewise, queries with more than 5 correct responses are omitted since the model already demonstrates competence in these cases and no further training is necessary.
In the end, we retain 66K queries for training.

\paragraph{Group Relative Policy Optimization~(GRPO).}

As introduced by \cite{deepseekmath}, GRPO is a reinforcement learning method specifically designed for large language models,  obviating the need for additional value function approximation as in PPO. GRPO uses the average rewards of a group of sampled outputs as a baseline to calculate the advantages of each output. The objective of GRPO is defined as Eq. \ref{eq:grpo}:
{\footnotesize
\begin{align}\label{eq:grpo}
    \mathcal{J}_{GRPO}(\theta) &= \mathbb{E}_{[q\sim P(Q),\{o_i\}_{i=1}^G\sim \pi_{\theta_{old}}(O|q)]} \notag \\
    &\frac{1}{G}\sum_{i=1}^G\frac{1}{\vert o_i\vert}\sum_{t=1}^{\vert o_i\vert}\{\min[\frac{\pi_{\theta}^{i,t}}{\pi_{\theta_{old}}^{i,t}}\hat{A}_{i,t},\textrm{clip}(\frac{\pi_{\theta}^{i,t}}{\pi_{\theta_{old}}^{i,t}}, 1-\epsilon, 1+\epsilon)\hat{A}_{i,t}]-\beta\mathbb{D}_{KL}[\pi_{\theta}\vert\vert\pi_{\textrm{ref}}]\},
\end{align}
}
where $\pi^{i,t}=\pi(o_{i,t}|q, o_{i,<t})$, G is the number of responses in a group. $\pi_{ref}$, $\pi_{\theta}$, and $\pi_{old}$ are reference, training, and sampling models, respectively. $q$ and $\{o_i\}_{i=1}^G$ are questions and generated responses set in training. The advantage of each responses $\hat{A}_i$ is calculated by $\hat{A}_i=\frac{r_i - \text{mean}(r_i)}{\text{std}(r_i)}$. Then this sequence-level advantage is applied to each token in the response as $\hat{A}_{i,t}$.

\paragraph{Reward Shaping.}
We combine the rewards from both a rule-based verifier and the reward model to shape the overall reward signal.
The rule-based verifier extracts potential answers from each response and compares them against the gold-standard answer. 
 
Given that the output of the reward model is denoted as $r_m\in \mathbb{R}$, 
and the sparse reward from the rule-based verifier as $r_v\in\{0, 1\}$, the overall reward is calculated as follows:
\begin{equation}\label{eq:reward}
    r= \sigma(\alpha \cdot r_m) + (r_v - 1),
\end{equation}
where $\alpha$ is set as $0.5$ in all of our experiments.

This shaping mechanism ensures that correct responses consistently receive higher overall rewards compared to incorrect ones. 
Within each of the correct and incorrect groups, the responses are ranked based on the scores from the reward models.
ecially in hard samples.

\paragraph{Implementations.}
Our experiments are implemented based on the open-source RLHF framework ChatLearn\footnote{\url{https://github.com/alibaba/ChatLearn}}. The core implementation of our rule-based verifier is similar to the one used in our evaluation\footnote{\url{https://github.com/QwenLM/Qwen2-Math/tree/main/evaluation}}. All policy models in different parameter sizes are trained with the same reward model. We sample 32 responses for each query. Considering a pair of queries and responses as a sample, the number of samples in one episode is 4,096 and 2,048 for training 7B and 72B, respectively. All models are trained with a 512 global batch size. The learning rates are $1\times 10^{-5}$ and $5\times 10^{-6}$ for 7B and 72B, respectively.
And the KL coefficient for all training is $1\times 10^{-3}$. We mask all output tokens the Python executor provides in reinforcement learning of tool-integrated reasoning.

\section{Decontamination}
Decontamination is critical to ensuring unbiased model performance evaluation. 
Following prior work~\citep{qwen2}, we exclude potentially contaminated training samples using 13-gram matching. 
To improve the accuracy of this matching process, we perform text normalization, removing irrelevant punctuation and symbols.
To further reduce false negatives, particularly for common mathematical expressions, we introduce an additional criterion: the ratio of the longest common subsequence must exceed $0.6$ for a sample to be considered contaminated. 
For pre-training data, we filter potentially contaminated samples against datasets such as GSM8K~\citep{gsm8k} and MATH~\citep{math}.
When dealing with post-training data, including SFT data, RM training data, and the RL query set, we exclude any potentially contaminated problems or solutions across all reported evaluation datasets. These evaluation datasets include GSM8K~\citep{gsm8k}, MATH~\citep{math}, Minerva Math~\citep{minerva}, Gaokao 2023 En~\citep{mario}, Olympiad Bench~\citep{OlympiadBench}, College Math~\citep{mathscale}, MMLU STEM~\citep{mmlu}, GaoKao~\citep{agieval}, CMATH~\citep{cmath}, CN Middle School 24, AIME 24, and AMC 23.
During the analysis of contaminated samples, we identify that some existing training datasets (e.g., the MATH training dataset) contain a significant proportion of problems that share highly similar concepts or structures with those found in test datasets. 
Although these variations are not exact duplicates, they could potentially compromise the integrity of our evaluation. 
Therefore, we continue to exclude such samples from the training corpora. Table~\ref{tab:problem-comparison} provides examples of similar problems identified across the training and test sets.

\begin{table}[htbp]
\centering
\begin{tabular}{p{0.45\textwidth} p{0.45\textwidth}}
\textbf{Problems from MATH train (filtered):} & \textbf{Problems from MATH test:} \\[0.5em]
What is the remainder when $1 + 2 + 3 + 4 + \dots + 9 + 10$ is  divided by 8? & What is the remainder when $1 + 2 + 3 + 4 + \dots + 9 + 10$ is divided by 9? \\
\\
For how many integer values of $n$ between 1 and 1000 inclusive does the decimal representation of $\frac{n}{1400}$ terminate? & For how many integer values of $n$ between 1 and 1000 inclusive does the decimal representation of $\frac{n}{1375}$ terminate? \\
\\
Krista put 1 cent into her new bank on a Sunday morning.  On Monday she put 2 cents into her bank.  On Tuesday she put 4 cents into her bank, and she continued to double the amount of money she put into her bank each day for two weeks.  On what day of the week did the total amount of money in her bank first exceed $\$2$? & Krista put 1 cent into her new bank on a Sunday morning.  On Monday she put 2 cents into her bank.  On Tuesday she put 4 cents into her bank, and she continued to double the amount of money she put into her bank each day for two weeks.  On what day of the week did the total amount of money in her bank first exceed $\$5$? \\
\end{tabular}
\caption{Examples of filtered samples in the MATH training set with similar samples in the test set.}
\label{tab:problem-comparison}
\end{table}

\section{Evaluation}
\label{sec:experiment}

\subsection{Base Models}

We evaluate our Qwen2-Math and Qwen2.5-Math base models on three widely used English math benchmarks GSM8K~\citep{gsm8k}, MATH~\citep{math}, and MMLU-STEM~\citep{mmlu}. In addition, we also evaluate three Chinese math benchmarks CMATH~\citep{cmath}, GaoKao Math Cloze~\citep{agieval}, and GaoKao Math QA~\citep{agieval}. All evaluations are tested with few-shot chain-of-thought prompting. The prompts of these benchmarks are shown in Appendix~\ref{sec:prompts}. For general models, we report the results on LLama-3.1-8B/70B/405B~\citep{llama3} and Qwen2-1.5B/7B/72B~\citep{qwen2}. For specific models, DeepSeekMath-Base-7B~\citep{deepseekmath}, DeepSeek-Coder-V2-Lite-Base~\citep{deepseekcoderv2}, and Intermln2-Math-Base-20B~\citep{internlmmath} are used as baselines.

The results are shown in Table~\ref{tab:base}. We can see that the smallest model of the Qwen2.5-Math series, Qwen2.5-Math-1.5B, outperforms all specific baselines on GSM8K, MATH, CMATH, GaoKao Math Cloze, and Gaokao Math QA. Furthermore, the medium-size model, Qwen2.5-Math-7B, obtains 91.6 and 55.4 scores on GSM8K and MATH, which outperforms Qwen2-72B with 89.5 and 51.1, and Llama-3.1-405B with 89.0 and 53.8. Our flagship Qwen2.5-Math-72B achieves new SOTA on MATH, CMATH, Gaokao Math Cloze, and Gaokao Math QA, which obtains 66.8 on MATH. Compared to Qwen2-Math-1.5B/7B/72B, Qwen2.5-Math-1.5B/7B/72B have achieved significant improvements on all benchmarks. For example, Qwen2.5-Math-1.5B/7B/72B obtains 5.4, 5.0, 6.3 scores improvement on MATH, and 3.4, 12.2, 19.8 scores improvement on Gaokao Math QA, which demonstrates the effectiveness of our \textit{Qwen Math corpus v2}.

\begin{table}[htbp]
    \small
   \centering
   \scalebox{0.9}{
    \begin{tabular}{l|ccc|ccc}
    \toprule
    \multicolumn{1}{c|}{\multirow{2}[4]{*}{\diagbox{\textbf{\textsc{Model}}}{\textbf{\textsc{Benchmark}}}}} & \multicolumn{3}{c}{\textbf{\textsc{En}}} & \multicolumn{3}{c}{\textbf{\textsc{Zh}}} \\
\cmidrule{2-7} & \textbf{GSM8K} & \textbf{MATH}  & \textbf{\makecell{MMLU\\STEM}} & \textbf{CMATH} & \textbf{\makecell{GaoKao\\Math Cloze}} & \textbf{\makecell{GaoKao\\Math QA}} \\
        & \textit{8-shot} & \textit{4-shot} & \textit{4-shot} & \textit{6-shot} & \textit{5-shot} & \textit{4-shot} \\
    \midrule
    \multicolumn{7}{c}{\small \textit{General Model}} \\
    \midrule
    Llama-3.1-8B & 56.7  & 20.3  & 53.1  & 51.5  & 8.5   & 28.5  \\
    Llama-3.1-70B & 85.5  & 41.4  & 78.1  & 75.5  & 11.9  & 43.3  \\
    Llama-3.1-405B & 89.0  & 53.8  & -     & -     & -     & - \\
    Qwen2-1.5B & 58.5  & 21.7  & 44.8  & 55.6  & 12.7  & 35.6  \\
    Qwen2-7B & 79.9  & 44.2  & 67.6  & 76.7  & 37.3  & 51.6  \\
    Qwen2-72B & 89.5  & 51.1  & 79.9  & 85.4  & 55.9  & 72.6  \\
    \midrule
    \multicolumn{7}{c}{\small \textit{Specific Model}} \\
    \midrule
    DeepSeekMath-Base-7B & 64.2  & 36.2  & 56.5  & 71.7  & 20.3  & 40.7  \\
    DeepSeek-Coder-V2-Lite-Base & 68.3  & 38.1  & 59.5  & 77.8  & 25.4  & 51.3  \\
    Internlm2-Math-Base-20B & 68.2  & 30.4  & 63.0  & 65.9  & 16.9  & 40.2  \\
    Qwen2-Math-1.5B & 71.3  & 44.4  & 50.4  & 79.6  & 37.3  & 50.7  \\
    Qwen2-Math-7B & 80.4  & 50.4  & 65.7  & 83.2  & 48.3  & 57.3  \\
    Qwen2-Math-72B & 89.1  & 60.5  & 79.1  & 86.4  & 72.9  & 69.5  \\
    \textbf{Qwen2.5-Math-1.5B} & 76.8  & 49.8  & 51.3  & 83.0  & 47.5  & 54.1  \\
     \textbf{Qwen2.5-Math-7B} & \textbf{91.6}  & 55.4  & 67.8  & 85.0  & 57.6  & 69.5  \\
    \textbf{Qwen2.5-Math-72B} & 90.8  & \textbf{66.8}  & \textbf{82.8}  & \textbf{89.7}  & \textbf{72.9}  & \textbf{86.3}  \\

    \bottomrule
    \end{tabular}}
  \caption{The results of Qwen2.5-Math and other base models on English and Chinese mathematical benchmarks. Models are evaluated with few-shot chain-of-thought prompting.}
  \label{tab:base}
\end{table}%

\begin{figure}[htbp]
    \centering
    \includegraphics[width=1.0\columnwidth]{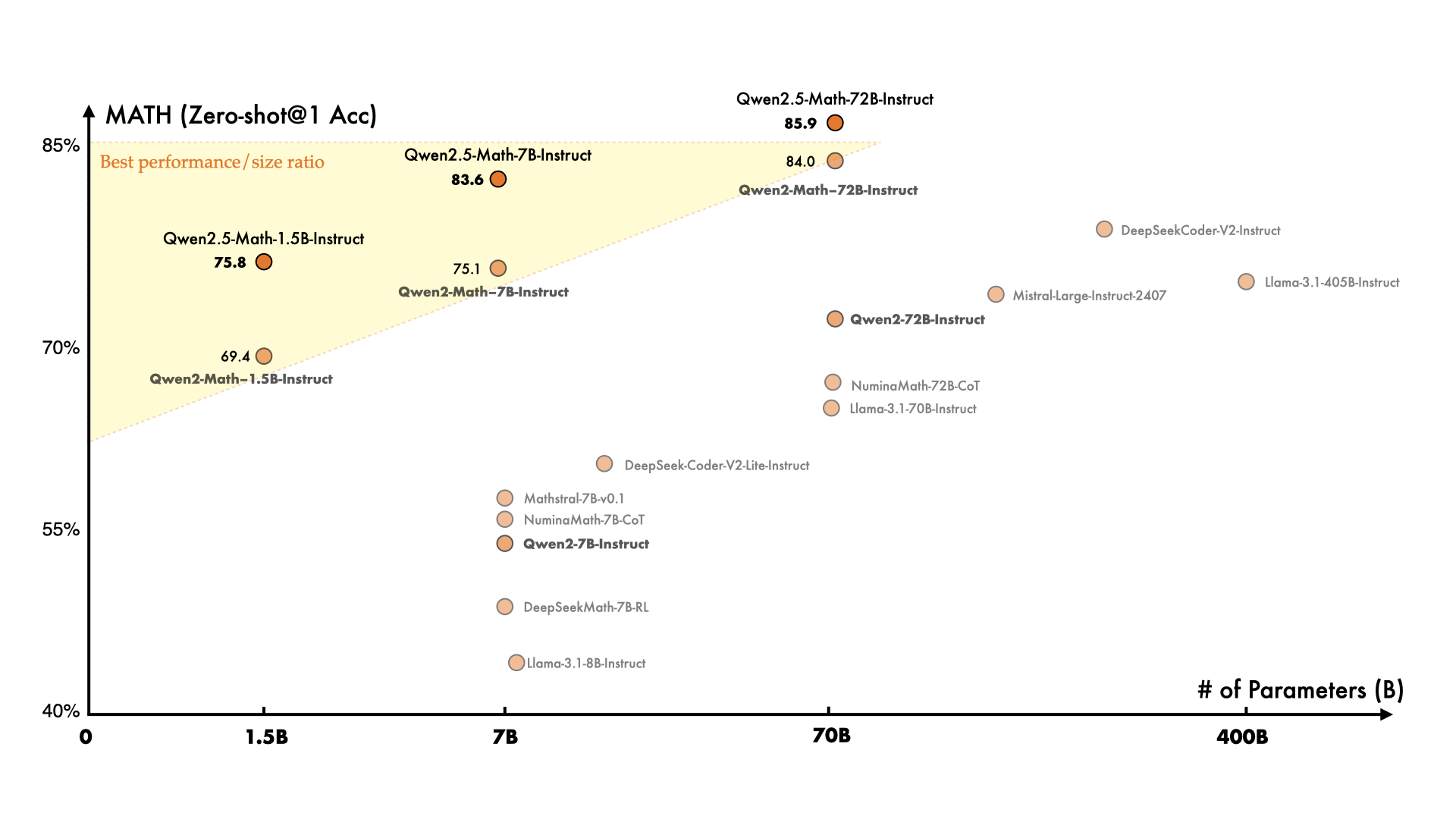}
    \vspace{-1mm}
    \caption{The Performance of Qwen2.5-Math-1.5/7/72B-Instruct on MATH by CoT compared to models of the same size.}
    \label{fig:exp}
\end{figure}
\subsection{Instruction Models}

We evaluate Qwen2-Math-Instruct on mathematical benchmarks in both English and Chinese. 
In addition to the widely-used benchmarks, such as GSM8K~\citep{gsm8k} and MATH~\citep{math}, we also involve more exams that are more challenging to fully inspect the capabilities of Qwen2-Math-Instruct and Qwen2.5-Math-Instruct, such as OlympiadBench~\citep{OlympiadBench}, CollegeMath~\citep{collegeMATH}, GaoKao 2023 En~\citep{mario}, AIME2024~\footnote{\url{https://huggingface.co/datasets/AI-MO/aimo-validation-aime}}, and AMC2023~\footnote{\url{https://huggingface.co/datasets/AI-MO/aimo-validation-amc}}. For Chinese mathematical benchmarks, we use CMATH~\citep{cmath}, GaoKao (including GaoKao I/II 2024~\footnote{\url{https://github.com/llmeval/Llmeval-Gaokao2024-Math}}, GaoKao-Math-QA~\citep{agieval}, GaoKao-Math-Cloze~\citep{agieval} and 91 collected GaoKao problems in 2024), and CN Middle School 24 (101 collected problems from China High School Entrance Examination in 2024).
We report greedy, Maj@8, and RM@8 performance on all benchmarks in the zero-shot setting, except for the multi-choice benchmarks (including MMLU STEM and multiple-choice problems in GaoKao and CN Middle School 24) with a 5-shot setting.

We take Qwen2-1.5/7/72B-Instruct~\citep{qwen2}, Llama-3.1-8/70B-instruct~\citep{llama3}, and GPT4o-2024-08-06~\citep{gpt4o} as general model baselines. Besides, DeepSeekMath-7B-RL~\citep{deepseekmath}, DeepSeek-Coder-V2-Lite-Instruct~\citep{deepseekcoderv2}, Interlm2-math-plus-7B/20B/mixtral8x7B~\citep{internlmmath}, Mathstral-7B-v0.1~\citep{mathstral}, NuminaMath-7/72B-CoT~\citep{numinamath} are taken as specific-model baselines.

\begin{table}[htbp]
  \centering
  \resizebox{\linewidth}{!}{
    \begin{tabular}{l|cccccccc}
    \toprule
    \multicolumn{1}{c|}{\multirow{2}[4]{*}{\diagbox{Model}{Benchmark}}} & \multicolumn{8}{c}{EN}\\
\cmidrule{2-9} & GSM8K & MATH  & \makecell{Minerva\\Math} & \makecell{GaoKao\\2023 En} & \makecell{Olympiad\\Bench} & \makecell{College\\Math} & \makecell{MMLU\\STEM} & \makecell{Avg.} \\
\midrule
    \multicolumn{9}{c}{\textbf{\textsc{Chain-of-Thought}}} \\ 
    \midrule
        GPT-4o-2024-08-06 & 92.9  & 81.1  & 36.8  & 67.5  & 43.3  & 48.5  & 64.2 & 62.0\\
    \midrule
    DeepSeekMath-7B-RL & 88.2  & 52.4  & 20.6  & 43.6  & 19.0  & 37.5  & 64.8 & 46.6  \\
    DeepSeek-Coder-V2-Lite-Instruct & 87.6  & 61.0  & 29.4  & 56.1  & 26.4  & 39.8  & 68.6 & 52.7   \\
    \midrule
    Internlm2-math-plus-7B & 84.0  & 54.4  & 17.3  & 50.1  & 18.8  & 36.2  & 55.2 & 45.1  \\
    Internlm2-math-plus-20B & 87.9  & 56.5  & 20.2  & 51.9  & 23.1  & 37.5  & 63.5 & 48.7 \\
    Internlm2-math-plus-mixtral8x7B & 92.1  & 59.4  & 26.8  & 49.6  & 25.0  & 37.5  & 71.9 & 51.8  \\
    \midrule
    Mathstral-7B-v0.1 & 84.9  & 56.6  & 16.2  & 46.0  & 21.5  & 33.7  & 64.0 & 46.1  \\
    \midrule
    NuminaMath-7B-CoT & 75.4  & 55.2  & 19.1  & 47.5  & 19.9  & 36.9  & 60.8 & 45.0   \\
    NuminaMath-72B-CoT & 90.8  & 66.7  & 25.0  & 58.4  & 32.6  & 39.7  & 64.5  & 54.0  \\
    \midrule
    Llama-3.1-8B-Instruct & 76.6  & 47.2  & 21.7  & 38.4  & 15.4  & 33.8  & 60.5 & 41.9    \\
    Llama-3.1-70B-Instruct & 94.1  & 65.7  & 34.2  & 54.0  & 27.7  & 42.5  & 80.4 & 56.9 \\
    \midrule
    Qwen2-1.5B-Instruct & 64.1  & 25.1  & 5.5   & 19.7  & 4.1   & 10.4  & 46.2  & 25.0 \\
    Qwen2-7B-Instruct & 85.7  & 52.9  & 19.5  & 36.4  & 21.3  & 24.5  & 68.2 & 44.1  \\
    Qwen2-72B-Instruct & 93.2  & 69.0  & 31.6  & 58.7  & 33.2  & 43.2  & \textbf{84.4} & 59.0  \\
    \midrule
    \multirow{2}[2]{*}{Qwen2-Math-1.5B-Instruct} & 84.2  & 69.4  & 29.4  & 59.7  & 31.3  & 44.2  & 54.9 & 53.3 \\
  & \hspace{0.63cm}88.6\textcolor[RGB]{10, 0, 15}{$_{\texttt{maj@8}}$}  & \hspace{0.63cm}75.3\textcolor[RGB]{10, 0, 15}{$_{\texttt{maj@8}}$}  & \hspace{0.63cm}32.0\textcolor[RGB]{10, 0, 15}{$_{\texttt{maj@8}}$}  & \hspace{0.63cm}63.9\textcolor[RGB]{10, 0, 15}{$_{\texttt{maj@8}}$}  & \hspace{0.63cm}37.6\textcolor[RGB]{10, 0, 15}{$_{\texttt{maj@8}}$}  & \hspace{0.63cm}46.6\textcolor[RGB]{10, 0, 15}{$_{\texttt{maj@8}}$}  & \hspace{0.63cm}59.5\textcolor[RGB]{10, 0, 15}{$_{\texttt{maj@8}}$}  & \hspace{0.63cm}57.6\textcolor[RGB]{10, 0, 15}{$_{\texttt{maj@8}}$} 
  \\
        & \hspace{0.5cm}92.7\textcolor[RGB]{84, 89, 247}{$_{\texttt{rm@8}}$}  & \hspace{0.5cm}79.9\textcolor[RGB]{84, 89, 247}{$_{\texttt{rm@8}}$}  & \hspace{0.5cm}36.4\textcolor[RGB]{84, 89, 247}{$_{\texttt{rm@8}}$}  & \hspace{0.5cm}68.8\textcolor[RGB]{84, 89, 247}{$_{\texttt{rm@8}}$}  & \hspace{0.5cm}43.4\textcolor[RGB]{84, 89, 247}{$_{\texttt{rm@8}}$}  & \hspace{0.5cm}46.8\textcolor[RGB]{84, 89, 247}{$_{\texttt{rm@8}}$}  & \hspace{0.5cm}74.5\textcolor[RGB]{84, 89, 247}{$_{\texttt{rm@8}}$}  & \hspace{0.5cm}63.2\textcolor[RGB]{84, 89, 247}{$_{\texttt{rm@8}}$}  \\
    \midrule
    \multirow{2}[2]{*}{Qwen2-Math-7B-Instruct} & 89.9  & 75.1  & 34.6  & 62.1  & 38.2  & 45.9  & 63.8 & 58.5   \\
      & \hspace{0.63cm}93.1\textcolor[RGB]{10, 0, 15}{$_{\texttt{maj@8}}$}  & \hspace{0.63cm}80.2\textcolor[RGB]{10, 0, 15}{$_{\texttt{maj@8}}$}  & \hspace{0.63cm}37.1\textcolor[RGB]{10, 0, 15}{$_{\texttt{maj@8}}$}  & \hspace{0.63cm}68.1\textcolor[RGB]{10, 0, 15}{$_{\texttt{maj@8}}$}  & \hspace{0.63cm}43.7\textcolor[RGB]{10, 0, 15}{$_{\texttt{maj@8}}$}  & \hspace{0.63cm}47.8\textcolor[RGB]{10, 0, 15}{$_{\texttt{maj@8}}$}  & \hspace{0.63cm}73.2\textcolor[RGB]{10, 0, 15}{$_{\texttt{maj@8}}$}  & \hspace{0.63cm}63.3\textcolor[RGB]{10, 0, 15}{$_{\texttt{maj@8}}$}   \\

        & \hspace{0.5cm}95.1\textcolor[RGB]{84, 89, 247}{$_{\texttt{rm@8}}$}  & \hspace{0.5cm}83.3\textcolor[RGB]{84, 89, 247}{$_{\texttt{rm@8}}$}  & \hspace{0.5cm}39.7\textcolor[RGB]{84, 89, 247}{$_{\texttt{rm@8}}$}  & \hspace{0.5cm}71.9\textcolor[RGB]{84, 89, 247}{$_{\texttt{rm@8}}$}  & \hspace{0.5cm}47.6\textcolor[RGB]{84, 89, 247}{$_{\texttt{rm@8}}$}  & \hspace{0.5cm}47.9\textcolor[RGB]{84, 89, 247}{$_{\texttt{rm@8}}$}  & \hspace{0.5cm}78.2\textcolor[RGB]{84, 89, 247}{$_{\texttt{rm@8}}$}  & \hspace{0.5cm}66.2\textcolor[RGB]{84, 89, 247}{$_{\texttt{rm@8}}$}   \\
    \midrule
    \multirow{2}[2]{*}{Qwen2-Math-72B-Instruct} & \textbf{96.7}  & 84.0  & 40.1  & 68.3  & 43.0  & 47.9  & 79.9 & 65.7  \\
          & \hspace{0.63cm}97.0\textcolor[RGB]{10, 0, 15}{$_{\texttt{maj@8}}$}  & \hspace{0.63cm}86.8\textcolor[RGB]{10, 0, 15}{$_{\texttt{maj@8}}$}  & \hspace{0.63cm}45.2\textcolor[RGB]{10, 0, 15}{$_{\texttt{maj@8}}$}  & \hspace{0.63cm}71.4\textcolor[RGB]{10, 0, 15}{$_{\texttt{maj@8}}$}  & \hspace{0.63cm}48.9\textcolor[RGB]{10, 0, 15}{$_{\texttt{maj@8}}$}  & \hspace{0.63cm}48.7\textcolor[RGB]{10, 0, 15}{$_{\texttt{maj@8}}$}  & \hspace{0.63cm}83.1\textcolor[RGB]{10, 0, 15}{$_{\texttt{maj@8}}$}  & \hspace{0.63cm}68.7\textcolor[RGB]{10, 0, 15}{$_{\texttt{maj@8}}$}  \\

        & \hspace{0.5cm}96.7\textcolor[RGB]{84, 89, 247}{$_{\texttt{rm@8}}$}  & \hspace{0.5cm}86.7\textcolor[RGB]{84, 89, 247}{$_{\texttt{rm@8}}$}  & \hspace{0.5cm}47.1\textcolor[RGB]{84, 89, 247}{$_{\texttt{rm@8}}$}  & \hspace{0.5cm}72.5\textcolor[RGB]{84, 89, 247}{$_{\texttt{rm@8}}$}  & \hspace{0.5cm}52.4\textcolor[RGB]{84, 89, 247}{$_{\texttt{rm@8}}$}  & \hspace{0.5cm}48.2\textcolor[RGB]{84, 89, 247}{$_{\texttt{rm@8}}$}  & \hspace{0.5cm}82.2\textcolor[RGB]{84, 89, 247}{$_{\texttt{rm@8}}$}  & \hspace{0.5cm}69.4\textcolor[RGB]{84, 89, 247}{$_{\texttt{rm@8}}$} \\
    \midrule
    \multirow{2}[2]{*}{\textbf{Qwen2.5-Math-1.5B-Instruct}} & 84.8  & 75.8  & 29.4  &  65.5 & 38.1  & 47.7  &  57.5 & 56.9  \\
  & \hspace{0.63cm}89.5\textcolor[RGB]{10, 0, 15}{$_{\texttt{maj@8}}$}  & \hspace{0.63cm}80.3\textcolor[RGB]{10, 0, 15}{$_{\texttt{maj@8}}$}  & \hspace{0.63cm}32.0\textcolor[RGB]{10, 0, 15}{$_{\texttt{maj@8}}$}  & \hspace{0.63cm}68.8\textcolor[RGB]{10, 0, 15}{$_{\texttt{maj@8}}$}  & \hspace{0.63cm}43.9\textcolor[RGB]{10, 0, 15}{$_{\texttt{maj@8}}$}  & \hspace{0.63cm}48.9\textcolor[RGB]{10, 0, 15}{$_{\texttt{maj@8}}$}  & \hspace{0.63cm}60.7\textcolor[RGB]{10, 0, 15}{$_{\texttt{maj@8}}$}  & \hspace{0.63cm}60.6\textcolor[RGB]{10, 0, 15}{$_{\texttt{maj@8}}$}  \\
        & \hspace{0.5cm}94.1\textcolor[RGB]{84, 89, 247}{$_{\texttt{rm@8}}$}  & \hspace{0.5cm}83.9\textcolor[RGB]{84, 89, 247}{$_{\texttt{rm@8}}$}  & \hspace{0.5cm}37.5\textcolor[RGB]{84, 89, 247}{$_{\texttt{rm@8}}$}  & \hspace{0.5cm}73.0\textcolor[RGB]{84, 89, 247}{$_{\texttt{rm@8}}$}  & \hspace{0.5cm}47.3\textcolor[RGB]{84, 89, 247}{$_{\texttt{rm@8}}$}  & \hspace{0.5cm}50.2\textcolor[RGB]{84, 89, 247}{$_{\texttt{rm@8}}$}  & \hspace{0.5cm}65.2\textcolor[RGB]{84, 89, 247}{$_{\texttt{rm@8}}$}  & \hspace{0.5cm}64.5\textcolor[RGB]{84, 89, 247}{$_{\texttt{rm@8}}$}   \\
    \midrule
    \multirow{2}[2]{*}{\textbf{Qwen2.5-Math-7B-Instruct}} &  95.2 & 83.6  & 37.1  & 66.8  &  41.6 & 46.8 & 71.9 & 62.9 \\
      & \hspace{0.63cm}96.7\textcolor[RGB]{10, 0, 15}{$_{\texttt{maj@8}}$}  & \hspace{0.63cm}87.1\textcolor[RGB]{10, 0, 15}{$_{\texttt{maj@8}}$}  & \hspace{0.63cm}41.2\textcolor[RGB]{10, 0, 15}{$_{\texttt{maj@8}}$}  & \hspace{0.63cm}72.5\textcolor[RGB]{10, 0, 15}{$_{\texttt{maj@8}}$}  & \hspace{0.63cm}44.4\textcolor[RGB]{10, 0, 15}{$_{\texttt{maj@8}}$}  & \hspace{0.63cm}47.8\textcolor[RGB]{10, 0, 15}{$_{\texttt{maj@8}}$}  & \hspace{0.63cm}73.8\textcolor[RGB]{10, 0, 15}{$_{\texttt{maj@8}}$}  & \hspace{0.63cm}66.2\textcolor[RGB]{10, 0, 15}{$_{\texttt{maj@8}}$}   \\

        & \hspace{0.5cm}97.9\textcolor[RGB]{84, 89, 247}{$_{\texttt{rm@8}}$}  & \hspace{0.5cm}88.5\textcolor[RGB]{84, 89, 247}{$_{\texttt{rm@8}}$}  & \hspace{0.5cm}42.6\textcolor[RGB]{84, 89, 247}{$_{\texttt{rm@8}}$}  & \hspace{0.5cm}75.1\textcolor[RGB]{84, 89, 247}{$_{\texttt{rm@8}}$}  & \hspace{0.5cm}49.9\textcolor[RGB]{84, 89, 247}{$_{\texttt{rm@8}}$}  & \hspace{0.5cm}49.6\textcolor[RGB]{84, 89, 247}{$_{\texttt{rm@8}}$}  & \hspace{0.5cm}78.7\textcolor[RGB]{84, 89, 247}{$_{\texttt{rm@8}}$}  & \hspace{0.5cm}68.9\textcolor[RGB]{84, 89, 247}{$_{\texttt{rm@8}}$} \\
    \midrule
    \multirow{2}[2]{*}{\textbf{Qwen2.5-Math-72B-Instruct}} & 95.9  & \textbf{85.9}  & \textbf{44.1}  & \textbf{71.9}  & \textbf{49.0}  & \textbf{49.5}  &  80.8 & \textbf{68.2}  \\
          & \hspace{0.63cm}96.0\textcolor[RGB]{10, 0, 15}{$_{\texttt{maj@8}}$}  & \hspace{0.63cm}88.6\textcolor[RGB]{10, 0, 15}{$_{\texttt{maj@8}}$}  & \hspace{0.63cm}47.8\textcolor[RGB]{10, 0, 15}{$_{\texttt{maj@8}}$}  & \hspace{0.63cm}73.8\textcolor[RGB]{10, 0, 15}{$_{\texttt{maj@8}}$}  & \hspace{0.63cm}50.1\textcolor[RGB]{10, 0, 15}{$_{\texttt{maj@8}}$}  & \hspace{0.63cm}50.2\textcolor[RGB]{10, 0, 15}{$_{\texttt{maj@8}}$}  & \hspace{0.63cm}84.9\textcolor[RGB]{10, 0, 15}{$_{\texttt{maj@8}}$}
          & \hspace{0.63cm}70.2\textcolor[RGB]{10, 0, 15}{$_{\texttt{maj@8}}$}
          \\

        & \hspace{0.5cm}96.4\textcolor[RGB]{84, 89, 247}{$_{\texttt{rm@8}}$}  & \hspace{0.5cm}89.8\textcolor[RGB]{84, 89, 247}{$_{\texttt{rm@8}}$}  & \hspace{0.5cm}47.4\textcolor[RGB]{84, 89, 247}{$_{\texttt{rm@8}}$}  & \hspace{0.5cm}76.9\textcolor[RGB]{84, 89, 247}{$_{\texttt{rm@8}}$}  & \hspace{0.5cm}54.5\textcolor[RGB]{84, 89, 247}{$_{\texttt{rm@8}}$} & \hspace{0.5cm}50.6\textcolor[RGB]{84, 89, 247}{$_{\texttt{rm@8}}$} & \hspace{0.5cm}80.1\textcolor[RGB]{84, 89, 247}{$_{\texttt{rm@8}}$} & \hspace{0.5cm}70.8\textcolor[RGB]{84, 89, 247}{$_{\texttt{rm@8}}$}\\
         \midrule 
            \multicolumn{9}{c}{\textbf{\textsc{Tool-integrated Reasoning}}}
    \\
    \midrule
    \multirow{2}[2]{*}{\textbf{Qwen2.5-Math-1.5B-Instruct}} & 83.7 & 79.9 & 33.5 & 67.8 & 49.2 & 54.8 & 56.9 & 60.8   \\
  & \hspace{0.63cm}90.0\textcolor[RGB]{10, 0, 15}{$_{\texttt{maj@8}}$}  & \hspace{0.63cm}85.3\textcolor[RGB]{10, 0, 15}{$_{\texttt{maj@8}}$}  & \hspace{0.63cm}35.3\textcolor[RGB]{10, 0, 15}{$_{\texttt{maj@8}}$}  & \hspace{0.63cm}71.9\textcolor[RGB]{10, 0, 15}{$_{\texttt{maj@8}}$}  & \hspace{0.63cm}54.3\textcolor[RGB]{10, 0, 15}{$_{\texttt{maj@8}}$}  & \hspace{0.63cm}56.3\textcolor[RGB]{10, 0, 15}{$_{\texttt{maj@8}}$}  & \hspace{0.63cm}60.4\textcolor[RGB]{10, 0, 15}{$_{\texttt{maj@8}}$}  & \hspace{0.63cm}64.8\textcolor[RGB]{10, 0, 15}{$_{\texttt{maj@8}}$}    \\
        & \hspace{0.5cm}93.3\textcolor[RGB]{84, 89, 247}{$_{\texttt{rm@8}}$}  & \hspace{0.5cm}88.9\textcolor[RGB]{84, 89, 247}{$_{\texttt{rm@8}}$}  & \hspace{0.5cm}39.7\textcolor[RGB]{84, 89, 247}{$_{\texttt{rm@8}}$}  & \hspace{0.5cm}78.7\textcolor[RGB]{84, 89, 247}{$_{\texttt{rm@8}}$}  & \hspace{0.5cm}59.3\textcolor[RGB]{84, 89, 247}{$_{\texttt{rm@8}}$}  & \hspace{0.5cm}58.8\textcolor[RGB]{84, 89, 247}{$_{\texttt{rm@8}}$}  & \hspace{0.5cm}76.6\textcolor[RGB]{84, 89, 247}{$_{\texttt{rm@8}}$}  & \hspace{0.5cm}70.8\textcolor[RGB]{84, 89, 247}{$_{\texttt{rm@8}}$}  \\
    \midrule
    \multirow{2}[2]{*}{\textbf{Qwen2.5-Math-7B-Instruct}} & 94.6 & 85.2 & 39.0 & 71.4 & 55.6 & 56.0 & 70.1 & 67.4   \\
      & \hspace{0.63cm}96.4\textcolor[RGB]{10, 0, 15}{$_{\texttt{maj@8}}$}  & \hspace{0.63cm}89.9\textcolor[RGB]{10, 0, 15}{$_{\texttt{maj@8}}$}  & \hspace{0.63cm}40.8\textcolor[RGB]{10, 0, 15}{$_{\texttt{maj@8}}$}  & \hspace{0.63cm}76.4\textcolor[RGB]{10, 0, 15}{$_{\texttt{maj@8}}$}  & \hspace{0.63cm}58.6\textcolor[RGB]{10, 0, 15}{$_{\texttt{maj@8}}$}  & \hspace{0.63cm}57.2\textcolor[RGB]{10, 0, 15}{$_{\texttt{maj@8}}$}  & \hspace{0.63cm}71.3\textcolor[RGB]{10, 0, 15}{$_{\texttt{maj@8}}$}  & \hspace{0.63cm}70.1\textcolor[RGB]{10, 0, 15}{$_{\texttt{maj@8}}$}  \\

        & \hspace{0.5cm}97.6\textcolor[RGB]{84, 89, 247}{$_{\texttt{rm@8}}$}  & \hspace{0.5cm}91.4\textcolor[RGB]{84, 89, 247}{$_{\texttt{rm@8}}$}  & \hspace{0.5cm}42.3\textcolor[RGB]{84, 89, 247}{$_{\texttt{rm@8}}$}  & \hspace{0.5cm}80.8\textcolor[RGB]{84, 89, 247}{$_{\texttt{rm@8}}$}  & \hspace{0.5cm}63.1\textcolor[RGB]{84, 89, 247}{$_{\texttt{rm@8}}$}  & \hspace{0.5cm}58.7\textcolor[RGB]{84, 89, 247}{$_{\texttt{rm@8}}$}  & \hspace{0.5cm}82.2\textcolor[RGB]{84, 89, 247}{$_{\texttt{rm@8}}$}  & \hspace{0.5cm}73.7\textcolor[RGB]{84, 89, 247}{$_{\texttt{rm@8}}$} \\
    \midrule
    \multirow{2}[2]{*}{\textbf{Qwen2.5-Math-72B-Instruct}} & \textbf{95.8} & \textbf{88.1} & \textbf{48.2} & \textbf{75.3} & \textbf{60.6} & \textbf{57.7} & \textbf{82.3} & \textbf{72.6}  \\
          & \hspace{0.63cm}96.7\textcolor[RGB]{10, 0, 15}{$_{\texttt{maj@8}}$}  & \hspace{0.63cm}91.8\textcolor[RGB]{10, 0, 15}{$_{\texttt{maj@8}}$}  & \hspace{0.63cm}48.2\textcolor[RGB]{10, 0, 15}{$_{\texttt{maj@8}}$}  & \hspace{0.63cm}83.1\textcolor[RGB]{10, 0, 15}{$_{\texttt{maj@8}}$}  & \hspace{0.63cm}64.5\textcolor[RGB]{10, 0, 15}{$_{\texttt{maj@8}}$}  & \hspace{0.63cm}58.3\textcolor[RGB]{10, 0, 15}{$_{\texttt{maj@8}}$}  & \hspace{0.63cm}85.0\textcolor[RGB]{10, 0, 15}{$_{\texttt{maj@8}}$}
          & \hspace{0.63cm}75.4\textcolor[RGB]{10, 0, 15}{$_{\texttt{maj@8}}$} \\

        & \hspace{0.5cm}96.4\textcolor[RGB]{84, 89, 247}{$_{\texttt{rm@8}}$}  & \hspace{0.5cm}92.9\textcolor[RGB]{84, 89, 247}{$_{\texttt{rm@8}}$}  & \hspace{0.5cm}49.3\textcolor[RGB]{84, 89, 247}{$_{\texttt{rm@8}}$}  & \hspace{0.5cm}83.4\textcolor[RGB]{84, 89, 247}{$_{\texttt{rm@8}}$}  & \hspace{0.5cm}65.9\textcolor[RGB]{84, 89, 247}{$_{\texttt{rm@8}}$}  & \hspace{0.5cm}59.7\textcolor[RGB]{84, 89, 247}{$_{\texttt{rm@8}}$}  & \hspace{0.5cm}90.0\textcolor[RGB]{84, 89, 247}{$_{\texttt{rm@8}}$}  & \hspace{0.5cm}76.8\textcolor[RGB]{84, 89, 247}{$_{\texttt{rm@8}}$}  \\
    \bottomrule
    \end{tabular}%

    }
     \caption{The results of Qwen2.5-Math-Instruct and other instruct models on English benchmarks. For CoT, we report few-shot pass@1 performance on MMLU(STEM) and zero-shot pass@1 performance on other benchmarks. For PoT, all benchmarks are evaluated in the zero-shot setting. Except for the pass@1 scores, we also provide the Qwen2-Math and Qwen2.5-Math performance with majority voting and reward model best-of-N among 8 sampled responses. Best pass@1 performance in CoT and TIR are marked in \textbf{bold}.}
\label{instruct_en_main_results}%

\end{table}%

\begin{table}[htbp]
  \centering
  \small
  \resizebox{0.7\linewidth}{!}{
    \begin{tabular}{l|cccc}
    \toprule
    \multicolumn{1}{c|}{\multirow{2}[4]{*}{\diagbox{Model}{Benchmark}}}   & \multicolumn{4}{c}{ZH} \\
\cmidrule{2-5} & GaoKao & CMATH & \makecell{CN Middle\\School 24} & \makecell{Avg.} \\
\midrule
    \multicolumn{5}{c}{\textbf{\textsc{Chain-of-Thought}}} \\ 
    \midrule
        GPT-4o-2024-08-06 & 42.6  & 92.5  &  60.4 & 65.2  \\
    \midrule
    DeepSeekMath-7B-RL & 33.6  & 86.7  & 67.3 & 62.5  \\
    DeepSeek-Coder-V2-Lite-Instruct & 51.1  & 89.8  & 66.3 & 69.1   \\
    \midrule
    Internlm2-math-plus-7B & 34.5  & 82.7  & 32.7 & 50.0  \\
    Internlm2-math-plus-20B & 36.1  & 81.3  & 33.7 & 50.4  \\
    Internlm2-math-plus-mixtral8x7B & 37.3  & 85.7  & 39.6 & 54.2  \\
    \midrule
    Mathstral-7B-v0.1 & 31.6  & 76.7  & 42.6 & 50.3  \\
    \midrule
    NuminaMath-7B-CoT & 36.4  & 78.2  & 60.4 & 58.3  \\
    NuminaMath-72B-CoT & 47.9  & 87.3  & 75.2 & 70.1  \\
    \midrule
    Llama-3.1-8B-Instruct & 30.4  & 64.8  & 43.6 & 46.3  \\
    Llama-3.1-70B-Instruct & 41.7  & 86.7  & 59.4 & 62.6  \\
    \midrule
    Qwen2-1.5B-Instruct & 17.0  & 65.5  & 31.7 & 38.1  \\
    Qwen2-7B-Instruct & 35.1  & 83.5  & 54.5 & 57.7  \\
    Qwen2-72B-Instruct & 54.6  & 92.2  & 74.3 & 73.7  \\
    \midrule
    \multirow{2}[2]{*}{Qwen2-Math-1.5B-Instruct} & 46.5  & 84.2  & 66.3 & 65.7  \\
   & \hspace{0.63cm}50.1\textcolor[RGB]{10, 0, 15}{$_{\texttt{maj@8}}$}  & \hspace{0.63cm}88.0\textcolor[RGB]{10, 0, 15}{$_{\texttt{maj@8}}$}  & \hspace{0.63cm}70.3\textcolor[RGB]{10, 0, 15}{$_{\texttt{maj@8}}$}  & \hspace{0.63cm}69.5\textcolor[RGB]{10, 0, 15}{$_{\texttt{maj@8}}$}  \\
        & \hspace{0.5cm}58.2\textcolor[RGB]{84, 89, 247}{$_{\texttt{rm@8}}$}  & \hspace{0.5cm}92.2\textcolor[RGB]{84, 89, 247}{$_{\texttt{rm@8}}$}  & \hspace{0.5cm}75.2\textcolor[RGB]{84, 89, 247}{$_{\texttt{rm@8}}$}  & \hspace{0.5cm}75.2\textcolor[RGB]{84, 89, 247}{$_{\texttt{rm@8}}$}  \\
    \midrule
    \multirow{2}[2]{*}{Qwen2-Math-7B-Instruct}  & 49.0  & 90.0  & 69.3 & 69.4   \\
       & \hspace{0.63cm}59.5\textcolor[RGB]{10, 0, 15}{$_{\texttt{maj@8}}$}  & \hspace{0.63cm}91.7\textcolor[RGB]{10, 0, 15}{$_{\texttt{maj@8}}$}  & \hspace{0.63cm}72.3\textcolor[RGB]{10, 0, 15}{$_{\texttt{maj@8}}$}  & \hspace{0.63cm}74.5 \textcolor[RGB]{10, 0, 15}{$_{\texttt{maj@8}}$}  \\

       & \hspace{0.5cm}62.7\textcolor[RGB]{84, 89, 247}{$_{\texttt{rm@8}}$}  & \hspace{0.5cm}94.0\textcolor[RGB]{84, 89, 247}{$_{\texttt{rm@8}}$}  & \hspace{0.5cm}78.2\textcolor[RGB]{84, 89, 247}{$_{\texttt{rm@8}}$}  & \hspace{0.5cm}78.3\textcolor[RGB]{84, 89, 247}{$_{\texttt{rm@8}}$}  \\

    \midrule
    \multirow{2}[2]{*}{Qwen2-Math-72B-Instruct} & 59.8   & 92.8   & 77.2  & 76.6    \\
    & \hspace{0.63cm}61.7\textcolor[RGB]{10, 0, 15}{$_{\texttt{maj@8}}$}  & \hspace{0.63cm}93.2\textcolor[RGB]{10, 0, 15}{$_{\texttt{maj@8}}$}  & \hspace{0.63cm}79.2\textcolor[RGB]{10, 0, 15}{$_{\texttt{maj@8}}$}  & \hspace{0.63cm}78.0 \textcolor[RGB]{10, 0, 15}{$_{\texttt{maj@8}}$}  \\

        & \hspace{0.5cm}67.7\textcolor[RGB]{84, 89, 247}{$_{\texttt{rm@8}}$}  & \hspace{0.5cm}94.2\textcolor[RGB]{84, 89, 247}{$_{\texttt{rm@8}}$}  & \hspace{0.5cm}78.2\textcolor[RGB]{84, 89, 247}{$_{\texttt{rm@8}}$}   & \hspace{0.5cm}80.0 \textcolor[RGB]{84, 89, 247}{$_{\texttt{rm@8}}$}  \\
    \midrule
    \multirow{2}[2]{*}{\textbf{Qwen2.5-Math-1.5B-Instruct}} &  62.4 & 89.7  & 76.2 & 76.1  \\
  & \hspace{0.63cm}66.4\textcolor[RGB]{10, 0, 15}{$_{\texttt{maj@8}}$}  & \hspace{0.63cm}91.7\textcolor[RGB]{10, 0, 15}{$_{\texttt{maj@8}}$}  & \hspace{0.63cm}77.2\textcolor[RGB]{10, 0, 15}{$_{\texttt{maj@8}}$}  & \hspace{0.63cm}78.4 \textcolor[RGB]{10, 0, 15}{$_{\texttt{maj@8}}$}  \\
       & \hspace{0.5cm}67.5\textcolor[RGB]{84, 89, 247}{$_{\texttt{rm@8}}$}  & \hspace{0.5cm}94.0\textcolor[RGB]{84, 89, 247}{$_{\texttt{rm@8}}$}  & \hspace{0.5cm}80.2\textcolor[RGB]{84, 89, 247}{$_{\texttt{rm@8}}$}  & \hspace{0.5cm}80.6 \textcolor[RGB]{84, 89, 247}{$_{\texttt{rm@8}}$}  \\
    \midrule
    \multirow{2}[2]{*}{\textbf{Qwen2.5-Math-7B-Instruct}} &  66.3  & 91.8  &  73.3 & 77.1  \\
       & \hspace{0.63cm}68.1\textcolor[RGB]{10, 0, 15}{$_{\texttt{maj@8}}$}  & \hspace{0.63cm}92.7\textcolor[RGB]{10, 0, 15}{$_{\texttt{maj@8}}$}  & \hspace{0.63cm}78.2\textcolor[RGB]{10, 0, 15}{$_{\texttt{maj@8}}$}  & \hspace{0.63cm}79.7 \textcolor[RGB]{10, 0, 15}{$_{\texttt{maj@8}}$}    \\

         & \hspace{0.5cm}72.2\textcolor[RGB]{84, 89, 247}{$_{\texttt{rm@8}}$}  & \hspace{0.5cm}94.5\textcolor[RGB]{84, 89, 247}{$_{\texttt{rm@8}}$}  & \hspace{0.5cm}81.2\textcolor[RGB]{84, 89, 247}{$_{\texttt{rm@8}}$}  & \hspace{0.5cm}82.6 \textcolor[RGB]{84, 89, 247}{$_{\texttt{rm@8}}$}  \\
    \midrule
    \multirow{2}[2]{*}{\textbf{Qwen2.5-Math-72B-Instruct}}  & \textbf{68.6}  & \textbf{94.3} & \textbf{79.2} & \textbf{82.7}   \\ & \hspace{0.63cm}75.0\textcolor[RGB]{10, 0, 15}{$_{\texttt{maj@8}}$}  & \hspace{0.63cm}95.3\textcolor[RGB]{10, 0, 15}{$_{\texttt{maj@8}}$}  & \hspace{0.63cm}79.2\textcolor[RGB]{10, 0, 15}{$_{\texttt{maj@8}}$}  & \hspace{0.63cm}83.2 \textcolor[RGB]{10, 0, 15}{$_{\texttt{maj@8}}$}  \\

         & \hspace{0.5cm}76.5\textcolor[RGB]{84, 89, 247}{$_{\texttt{rm@8}}$}  & \hspace{0.5cm}95.7\textcolor[RGB]{84, 89, 247}{$_{\texttt{rm@8}}$}  & \hspace{0.5cm}80.2\textcolor[RGB]{84, 89, 247}{$_{\texttt{rm@8}}$} & \hspace{0.63cm}84.1  \textcolor[RGB]{10, 0, 15}{$_{\texttt{rm@8}}$}  \\
         \midrule 
            \multicolumn{5}{c}{\textbf{\textsc{Tool-integrated Reasoning}}}
    \\
    \midrule
    \multirow{2}[2]{*}{\textbf{Qwen2.5-Math-1.5B-Instruct}}    & 59.6 & 89.3 & 71.3 & 73.4  \\
 & \hspace{0.63cm}68.3\textcolor[RGB]{10, 0, 15}{$_{\texttt{maj@8}}$}  & \hspace{0.63cm}90.8\textcolor[RGB]{10, 0, 15}{$_{\texttt{maj@8}}$}  & \hspace{0.63cm}78.2\textcolor[RGB]{10, 0, 15}{$_{\texttt{maj@8}}$} & \hspace{0.63cm}79.1 \textcolor[RGB]{10, 0, 15}{$_{\texttt{maj@8}}$} \\
         & \hspace{0.5cm}64.1\textcolor[RGB]{84, 89, 247}{$_{\texttt{rm@8}}$}  & \hspace{0.5cm}93.2\textcolor[RGB]{84, 89, 247}{$_{\texttt{rm@8}}$}  & \hspace{0.5cm}78.2\textcolor[RGB]{84, 89, 247}{$_{\texttt{rm@8}}$}  & \hspace{0.5cm}78.5  \textcolor[RGB]{84, 89, 247}{$_{\texttt{rm@8}}$}  \\
    \midrule
    \multirow{2}[2]{*}{\textbf{Qwen2.5-Math-7B-Instruct}} & 62.9 & 90.5 & 75.2 & 76.2   \\
       & \hspace{0.63cm}70.8\textcolor[RGB]{10, 0, 15}{$_{\texttt{maj@8}}$}  & \hspace{0.63cm}92.0\textcolor[RGB]{10, 0, 15}{$_{\texttt{maj@8}}$}  & \hspace{0.63cm}78.2\textcolor[RGB]{10, 0, 15}{$_{\texttt{maj@8}}$}  & \hspace{0.63cm}80.3 \textcolor[RGB]{10, 0, 15}{$_{\texttt{maj@8}}$}  \\

         & \hspace{0.5cm}72.9\textcolor[RGB]{84, 89, 247}{$_{\texttt{rm@8}}$}  & \hspace{0.5cm}94.2\textcolor[RGB]{84, 89, 247}{$_{\texttt{rm@8}}$}  & \hspace{0.5cm}80.2\textcolor[RGB]{84, 89, 247}{$_{\texttt{rm@8}}$}  & \hspace{0.5cm}82.4 \textcolor[RGB]{84, 89, 247}{$_{\texttt{rm@8}}$}  \\
    \midrule
    \multirow{2}[2]{*}{\textbf{Qwen2.5-Math-72B-Instruct}}  & \textbf{68.5} & \textbf{93.0} & \textbf{78.2} & \textbf{79.9}  \\
           & \hspace{0.63cm}72.0\textcolor[RGB]{10, 0, 15}{$_{\texttt{maj@8}}$}  & \hspace{0.63cm}93.5\textcolor[RGB]{10, 0, 15}{$_{\texttt{maj@8}}$}  & \hspace{0.63cm}78.2\textcolor[RGB]{10, 0, 15}{$_{\texttt{maj@8}}$} 
           & \hspace{0.63cm}81.2 \textcolor[RGB]{10, 0, 15}{$_{\texttt{maj@8}}$} \\

      & \hspace{0.5cm}75.4\textcolor[RGB]{84, 89, 247}{$_{\texttt{rm@8}}$}  & \hspace{0.5cm}93.8\textcolor[RGB]{84, 89, 247}{$_{\texttt{rm@8}}$}  & \hspace{0.5cm}80.2\textcolor[RGB]{84, 89, 247}{$_{\texttt{rm@8}}$}  & \hspace{0.5cm}83.1\textcolor[RGB]{84, 89, 247}{$_{\texttt{rm@8}}$} \\
    \bottomrule
    \end{tabular}%

    }
\caption{The results of Qwen2.5-Math-Instruct and other instruct models on Chinese benchmarks. }
\label{instruct_zh_main_results}%
\end{table}%

Let us first analyze the performance on English benchmarks. 
As shown in Table~\ref{instruct_en_main_results}, we can draw the following conclusions:
(1) Qwen2-Math-Instruct has demonstrated exceptional capabilities. The 1.5B model achieves an average score higher than any sub-70B model currently available. The 7B model performs on par with Qwen2-72B-Instruct, and Qwen2-Math-72B-Instruct surpasses the latest version of GPT-4o by 3.7 points.
(2) The performance of Qwen2.5-Math-Instruct represents a further upgrade over Qwen2-Math-Instruct. 
In the traditional CoT mode, the 1.5B and 7B Qwen2.5-Math-Instruct models achieve results comparable to the 7B and 72B Qwen2-Math-Instruct models, respectively, demonstrating a cross-scale improvement. 
Qwen2.5-Math-72B-Instruct achieves an average score of 2.5 points ahead of the current best model and is 6.2 points higher than GPT-4o. This shows that our improvements in training data and strategy can provide an alternative pathway for performance enhancements beyond simply increasing model size.
(3) The TIR mode introduced in Qwen2.5-Math-Instruct is highly effective. With the assistance of a Python Interpreter, the 7B model already matches the performance of Qwen2.5-Math-72B-Instruct. This indicates that precise mathematical calculations via external tools can significantly aid LLM reasoning. In many cases, the reasoning process of LLMs is sound, but computational errors can arise.
(4) Our RM performs exceptionally well. Across almost all benchmarks and models, RM@N scores are substantially better than Maj@N scores. This provides a reliable performance oracle for improving reinforcement learning strategies in the future. It is likely that we may soon see models with greedy decoding exceeding 90 points on MATH, even for the 7B scale.

\begin{figure}[htbp]
    \centering
    \includegraphics[width=0.8\columnwidth]{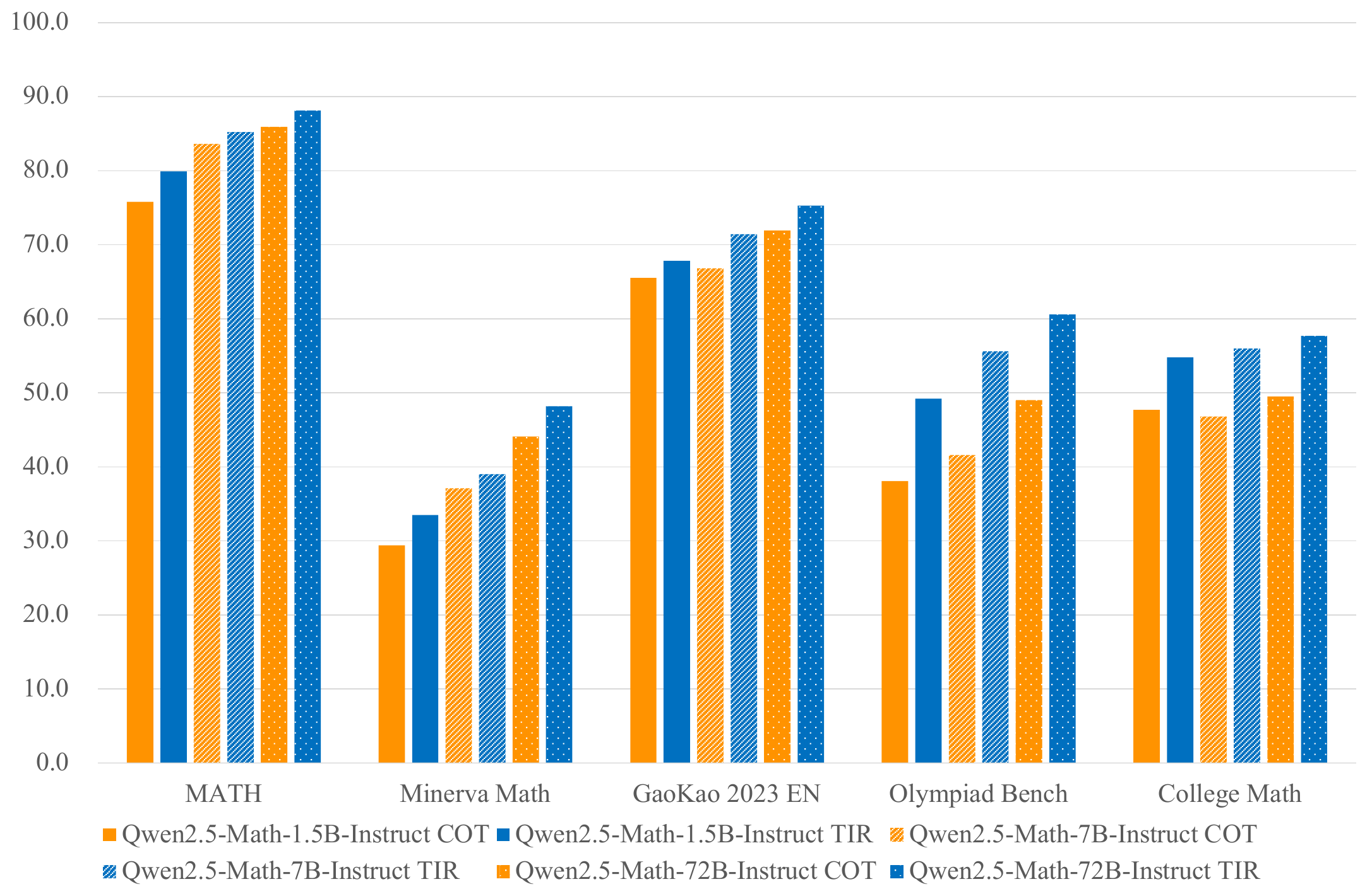}
    \vspace{-1mm}
    \caption{The Performance of Qwen2.5-Math-1.5/7/72B-Instruct by using TIR compared to using CoT. We use blue color to represent the performance of TIR, and orange to represent the performance of CoT. It can be seen that TIR can achieve further performance improvement compared to CoT.}
    \label{fig:exp_tir}
\end{figure}

Let's now shift our attention to Table~\ref{instruct_zh_main_results} to analyze the performance on the Chinese benchmarks.
For Qwen2-Math-Instruct, no specifically Chinese mathematics-related training data was incorporated. However, thanks to Qwen2's strong language transfer capabilities, the Qwen2-Math-1.5B-Instruct model has already surpassed GPT-4o in terms of the average Chinese score.
During the development of Qwen2.5-Math-Instruct, we intentionally integrated Chinese-specific math post-training data, resulting in substantial improvements in Chinese performance. 
The Qwen2.5-Math-1.5B-Instruct model achieves results similar to Qwen2-Math-72B-Instruct, while Qwen2.5-Math-72B-Instruct outperforms GPT-4o by an impressive 17.5 points.
Our RM also exhibits strong performance in Chinese benchmarks. Similar to our results in English, RM@N scores consistently surpass Maj@N scores, highlighting its effectiveness.
However, one key difference from the English results is that the TIR mode in Chinese does not show a significant performance advantage over the CoT mode. We will continue to investigate this aspect in future research.

Lastly, we intend to evaluate the model's ability to solve complex mathematical problems on highly challenging competition benchmarks such as AIME 2024 and AMC 2023.
As shown in Table~\ref{tab:math_competition}, we observe a significant improvement in performance on difficult problems with Qwen2.5-Math-Instruct compared to Qwen2-Math-Instruct.
With the support of the RM, Qwen2.5-Math-1.5B-Instruct, using the RM@256 in CoT mode, successfully solves 29 out of 40 problems on AMC 2023, significantly outperforming NuminaMath-72B CoT. 
Moreover, Qwen2.5-Math-72B-Instruct nearly achieves a perfect score in TIR mode, solving almost all the problems. 
We attribute this impressive performance to the extensive amounts of challenging mathematical data collected and synthesized during pre-training.
On the extremely difficult AIME 2024 benchmark, Claude3 Opus, GPT-4 Turbo, and Gemini 1.5 Pro manage to solve only 1 or 2 questions out of 30. 
In contrast, Qwen2.5-Math-72B-Instruct solves 9 problems in Greedy decoding CoT mode and 12 problems in TIR mode. With the help of the RM, Qwen2.5-Math-7B-Instruct could even solve up to 21 problems, further demonstrating the outstanding mathematical problem-solving ability of Qwen2.5-Math-Instruct.

\begin{table}[htbp]
  \centering
  \small
  \resizebox{0.7\linewidth}{!}{
    \begin{tabular}{l|cc}
    \toprule
    \multicolumn{1}{c|}{\textbf{\textsc{Model}}} & \multicolumn{1}{c}{\textbf{AIME24}} & \multicolumn{1}{c}{\textbf{AMC23}} \\
    \midrule
    \multicolumn{3}{c}{\textbf{\textsc{Chain-of-Thought}}} \\ 
    \midrule
    Claude 3 Opus & 2/30  & - \\
    GPT-4 Turbo & 1/30  & - \\
    Gemini 1.5 Pro & 2/30  & - \\
    \midrule
    \multirow{2}[0]{*}{Gemini Math-Specialized 1.5 Pro} & 7/30  & - \\
          & \hspace{0.8cm}8/30\textcolor[RGB]{84, 89, 247}{$_{\texttt{rm@256}}$}  & - \\
    \midrule
    \multirow{2}[1]{*}{NuminaMath-72B CoT} & 1/30  & 21/40 \\
          & \hspace{0.8cm}3/30\textcolor[RGB]{10, 0, 15}{$_{\texttt{maj@64}}$}  & \hspace{0.8cm}24/40\textcolor[RGB]{10, 0, 15}{$_{\texttt{maj@64}}$} \\
              \midrule
    \multirow{2}[0]{*}{Qwen2-Math-1.5B-Instruct} & 1/30  & 18/40 \\
          & \hspace{0.8cm}5/30\textcolor[RGB]{84, 89, 247}{$_{\texttt{rm@256}}$}  & \hspace{0.8cm}25/40\textcolor[RGB]{84, 89, 247}{$_{\texttt{rm@256}}$} \\
  \midrule
    \multirow{2}[0]{*}{Qwen2-Math-7B-Instruct} & 4/30  & 25/40 \\
          & \hspace{0.8cm}6/30\textcolor[RGB]{84, 89, 247}{$_{\texttt{rm@256}}$}  & \hspace{0.8cm}29/40\textcolor[RGB]{84, 89, 247}{$_{\texttt{rm@256}}$} \\
  \midrule
    \multirow{4}[1]{*}{Qwen2-Math-72B-Instruct} & 6/30  & 24/40 \\
              & \hspace{0.8cm}8/30\textcolor[RGB]{10, 0, 15}{$_{\texttt{maj@64}}$} & \hspace{0.8cm}29/40\textcolor[RGB]{10, 0, 15}{$_{\texttt{maj@64}}$} \\
          & \hspace{0.7cm}9/30\textcolor[RGB]{84, 89, 247}{$_{\texttt{rm@64}}$} & \hspace{0.7cm}29/40\textcolor[RGB]{84, 89, 247}{$_{\texttt{rm@64}}$} \\
          & \hspace{0.9cm}11/30\textcolor[RGB]{84, 89, 247}{$_{\texttt{rm@256}}$} & \hspace{0.8cm}28/40\textcolor[RGB]{84, 89, 247}{$_{\texttt{rm@256}}$} \\
      \midrule  
    \multirow{2}[0]{*}{Qwen2.5-Math-1.5B-Instruct} &  3/30 & 24/40 \\
          & \hspace{0.8cm}10/30\textcolor[RGB]{84, 89, 247}{$_{\texttt{rm@256}}$}  & \hspace{0.8cm}29/40\textcolor[RGB]{84, 89, 247}{$_{\texttt{rm@256}}$} \\
  \midrule
    \multirow{2}[0]{*}{Qwen2.5-Math-7B-Instruct} & 5/30 & 25/40 \\
          & \hspace{0.8cm}10/30\textcolor[RGB]{84, 89, 247}{$_{\texttt{rm@256}}$}  & \hspace{0.8cm}30/40\textcolor[RGB]{84, 89, 247}{$_{\texttt{rm@256}}$} \\
  \midrule
    \multirow{4}[1]{*}{Qwen2.5-Math-72B-Instruct} & 9/30 & 28/40 \\
              & \hspace{0.8cm}9/30\textcolor[RGB]{10, 0, 15}{$_{\texttt{maj@64}}$} & \hspace{0.8cm}\textbf{30/40}\textcolor[RGB]{10, 0, 15}{$_{\texttt{maj@64}}$} \\
          & \hspace{0.7cm}\textbf{13/30}\textcolor[RGB]{84, 89, 247}{$_{\texttt{rm@64}}$} & \hspace{0.7cm}29/40\textcolor[RGB]{84, 89, 247}{$_{\texttt{rm@64}}$} \\
          & \hspace{0.8cm}13/30\textcolor[RGB]{84, 89, 247}{$_{\texttt{rm@256}}$} & \hspace{0.8cm}30/40\textcolor[RGB]{84, 89, 247}{$_{\texttt{rm@256}}$} \\
\midrule
            \multicolumn{3}{c}{\textbf{\textsc{Tool-integrated Reasoning}}} \\ 
    \midrule  
    \multirow{2}[0]{*}{Qwen2.5-Math-1.5B-Instruct} & 7/30 & 20/40 \\
          & \hspace{0.8cm}9/30\textcolor[RGB]{10, 0, 15}{$_{\texttt{maj@64}}$}  & \hspace{0.8cm}31/40\textcolor[RGB]{10, 0, 15}{$_{\texttt{maj@64}}$} \\
          & \hspace{0.8cm}18/30\textcolor[RGB]{84, 89, 247}{$_{\texttt{rm@64}}$}  & \hspace{0.65cm}36/40\textcolor[RGB]{84, 89, 247}{$_{\texttt{rm@64}}$} \\
          & \hspace{0.9cm}9/30\textcolor[RGB]{10, 0, 15}{$_{\texttt{maj@256}}$}  & \hspace{0.9cm}32/40\textcolor[RGB]{10, 0, 15}{$_{\texttt{maj@256}}$} \\
          & \hspace{0.9cm}19/30\textcolor[RGB]{84, 89, 247}{$_{\texttt{rm@256}}$}  & \hspace{0.8cm}36/40\textcolor[RGB]{84, 89, 247}{$_{\texttt{rm@256}}$} \\
          
  \midrule
    \multirow{2}[0]{*}{Qwen2.5-Math-7B-Instruct} & 6/30 & 27/40 \\
    & \hspace{0.9cm}13/30\textcolor[RGB]{10, 0, 15}{$_{\texttt{maj@64}}$}  & \hspace{0.8cm}31/40\textcolor[RGB]{10, 0, 15}{$_{\texttt{maj@64}}$} \\
          & \hspace{0.8cm}\textbf{21/30}\textcolor[RGB]{84, 89, 247}{$_{\texttt{rm@64}}$}  & \hspace{0.7cm}33/40\textcolor[RGB]{84, 89, 247}{$_{\texttt{rm@64}}$} \\
    & \hspace{1.0cm}14/30\textcolor[RGB]{10, 0, 15}{$_{\texttt{maj@256}}$}  & \hspace{0.9cm}31/40\textcolor[RGB]{10, 0, 15}{$_{\texttt{maj@256}}$} \\
          & \hspace{0.9cm}21/30\textcolor[RGB]{84, 89, 247}{$_{\texttt{rm@256}}$}  & \hspace{0.8cm}35/40\textcolor[RGB]{84, 89, 247}{$_{\texttt{rm@256}}$} \\
  \midrule
    \multirow{4}[1]{*}{Qwen2.5-Math-72B-Instruct} & 12/30 & 28/40 \\
              & \hspace{0.8cm}14/30\textcolor[RGB]{10, 0, 15}{$_{\texttt{maj@64}}$} & \hspace{0.7cm}36/40\textcolor[RGB]{10, 0, 15}{$_{\texttt{maj@64}}$} \\
          & \hspace{0.7cm}18/30\textcolor[RGB]{84, 89, 247}{$_{\texttt{rm@64}}$} & \hspace{0.65cm}37/40\textcolor[RGB]{84, 89, 247}{$_{\texttt{rm@64}}$} \\
          & \hspace{0.9cm}16/30\textcolor[RGB]{10, 0, 15}{$_{\texttt{maj@256}}$} & \hspace{0.8cm}36/40\textcolor[RGB]{10, 0, 15}{$_{\texttt{maj@256}}$} \\
          & \hspace{0.8cm}19/30\textcolor[RGB]{84, 89, 247}{$_{\texttt{rm@256}}$} & \hspace{0.7cm}\textbf{39/40}\textcolor[RGB]{84, 89, 247}{$_{\texttt{rm@256}}$} \\
    \bottomrule
    \end{tabular}%
    }
 
\caption{The results on the mathematics competition problems. }
 \label{tab:math_competition}%
\end{table}%

\section{Conclusion}
\label{sec:conclusion}

In this report, we introduce Qwen2.5-Math, which features several key technical highlights: 
(1) extensive use of synthesized mathematical data from Qwen2-Math during the pre-training phase, 
(2) iterative generation of fine-tuning data and reinforcement training guided by the reward model during the post-training and inference phase and 
(3) support for bilingual (English and Chinese) queries, along with chain-of-thought and tool-integrated reasoning capabilities. 
As a result, Qwen2.5-Math represents the most advanced open-source math model series to date.
The Qwen2.5-Math-1.5B-Instruct model already surpasses most previous 70B math models, while the Qwen2.5-Math-7B-Instruct matches the performance of Qwen2-Math-72B-Instruct. 
Our flagship model, Qwen2.5-Math-7B-Instruct, outperforms Qwen2-Math-72B-Instruct with an average score increase of 4.4 points across 7 datasets.
We hope that the advances we’ve made with specialized models like Qwen2.5-Math will continue to strengthen the overall capabilities of the Qwen model and bring us closer to achieving artificial general intelligence.

\section*{Acknowledgements}

We sincerely appreciate the support from other members of the Qwen team. We would also like to thank the ChatLearn team from PAI, Alibaba, for their infrastructure support of large-scale reinforcement learning.

\clearpage

\bibliography{biblio}
\bibliographystyle{iclr2023_conference}
\clearpage

\appendix

\section{Case Study of Qwen2-MATH on Olympiad-level Problems}
\label{sec:appendix}
\subsection{Number Theory}
\noindent\textbf{Problem From IMO Shortlist 2002:}

What is the smallest positive integer $t$ such that there exist integers $x_1,x_2,\ldots,x_t$ with  \[x^3_1+x^3_2+\,\ldots\,+x^3_t=2002^{2002}\,?\]

\noindent\textbf{Qwen2-Math-72B-Instruct Solution}

To determine the smallest positive integer \( t \) such that there exist integers \( x_1, x_2, \ldots, x_t \) satisfying the equation

\[
x_1^3 + x_2^3 + \cdots + x_t^3 = 2002^{2002},
\]

we start by analyzing the properties of cubes and their sums. We need to find the smallest \( t \) for which the sum of \( t \) cubes equals \( 2002^{2002} \).

First, we note that \( 2002 \) can be factored into prime factors:

\[
2002 = 2 \times 7 \times 11 \times 13.
\]

Thus,

\[
2002^{2002} = (2 \times 7 \times 11 \times 13)^{2002} = 2^{2002} \times 7^{2002} \times 11^{2002} \times 13^{2002}.
\]

Next, we consider the properties of cubes modulo 9. Cubes modulo 9 can only be congruent to \( 0, 1, \) or \( -1 \). This is because:

\[
\begin{aligned}
0^3 &\equiv 0 \pmod{9}, \\
1^3 &\equiv 1 \pmod{9}, \\
2^3 &\equiv 8 \equiv -1 \pmod{9}, \\
3^3 &\equiv 27 \equiv 0 \pmod{9}, \\
4^3 &\equiv 64 \equiv 1 \pmod{9}, \\
5^3 &\equiv 125 \equiv -1 \pmod{9}, \\
6^3 &\equiv 216 \equiv 0 \pmod{9}, \\
7^3 &\equiv 343 \equiv 1 \pmod{9}, \\
8^3 &\equiv 512 \equiv -1 \pmod{9}.
\end{aligned}
\]

Now, we need to determine \( 2002^{2002} \mod 9 \). First, we find \( 2002 \mod 9 \):

\[
2002 \div 9 = 222 \text{ remainder } 4 \quad \Rightarrow \quad 2002 \equiv 4 \pmod{9}.
\]

Thus,

\[
2002^{2002} \equiv 4^{2002} \pmod{9}.
\]

Next, we use Euler's theorem, which states that \( a^{\phi(n)} \equiv 1 \pmod{n} \) for \( \gcd(a, n) = 1 \). Here, \( \phi(9) = 6 \), so:

\[
4^6 \equiv 1 \pmod{9}.
\]

We can reduce the exponent \( 2002 \mod 6 \):

\[
2002 \div 6 = 333 \text{ remainder } 4 \quad \Rightarrow \quad 2002 \equiv 4 \pmod{6}.
\]

Thus,

\[
4^{2002} \equiv 4^4 \pmod{9}.
\]

\[
4^4 = 256.
\]

Finally, we find \( 256 \mod 9 \):

\[
256 \div 9 = 28 \text{ remainder } 4 \quad \Rightarrow \quad 256 \equiv 4 \pmod{9}.
\]

Therefore,

\[
2002^{2002} \equiv 4 \pmod{9}.
\]

We need to find the smallest \( t \) such that the sum of \( t \) cubes is congruent to 4 modulo 9. The possible sums of cubes modulo 9 are:

\[
\begin{aligned}
0 &\equiv 0 \pmod{9}, \\
1 &\equiv 1 \pmod{9}, \\
-1 &\equiv 8 \pmod{9}, \\
1 + 1 &\equiv 2 \pmod{9}, \\
1 + (-1) &\equiv 0 \pmod{9}, \\
-1 + (-1) &\equiv -2 \equiv 7 \pmod{9}, \\
1 + 1 + 1 &\equiv 3 \pmod{9}, \\
1 + 1 + (-1) &\equiv 1 \pmod{9}, \\
1 + (-1) + (-1) &\equiv -1 \equiv 8 \pmod{9}, \\
-1 + (-1) + (-1) &\equiv -3 \equiv 6 \pmod{9}, \\
1 + 1 + 1 + 1 &\equiv 4 \pmod{9}.
\end{aligned}
\]

We see that the sum of four cubes can be congruent to 4 modulo 9. Therefore, the smallest \( t \) is 4.

Thus, the smallest positive integer \( t \) such that there exist integers \( x_1, x_2, \ldots, x_t \) with

\[
x_1^3 + x_2^3 + \cdots + x_t^3 = 2002^{2002}
\]

is

\[
\boxed{4}.
\]

\noindent\textbf{Problem From IMO Shortlist 2022}

For a positive integer $n$ we denote by $s(n)$ the sum of the digits of $n$. Let $P(x)=x^n+a_{n-1}x^{n-1}+\cdots+a_1x+a_0$ be a polynomial, where $n \geq 2$ and $a_i$ is a positive integer for all $0 \leq i \leq n-1$. Could it be the case that, for all positive integers $k$, $s(k)$ and $s(P(k))$ have the same parity?

\noindent\textbf{Qwen2-Math-72B-Instruct Solution}

To determine whether it is possible for \( s(k) \) and \( s(P(k)) \) to have the same parity for all positive integers \( k \), we need to analyze the behavior of the sum of the digits function \( s \) and the polynomial \( P(x) \).

First, recall that the sum of the digits of a number \( k \) modulo 2 is equivalent to \( k \mod 2 \). This is because the sum of the digits of \( k \) in base 10 is congruent to \( k \mod 9 \), and since \( 9 \equiv 1 \mod 2 \), the sum of the digits modulo 2 is the same as \( k \mod 2 \).

Given the polynomial \( P(x) = x^n + a_{n-1}x^{n-1} + \cdots + a_1x + a_0 \), we need to determine the parity of \( P(k) \) for any positive integer \( k \). Specifically, we need to check if \( P(k) \mod 2 \) is the same as \( k \mod 2 \).

Let's consider the polynomial modulo 2:
\[ P(x) \equiv x^n + a_{n-1}x^{n-1} + \cdots + a_1x + a_0 \mod 2. \]

Since \( a_i \) are positive integers, \( a_i \mod 2 \) is either 0 or 1. Let \( b_i = a_i \mod 2 \). Then we can rewrite the polynomial modulo 2 as:
\[ P(x) \equiv x^n + b_{n-1}x^{n-1} + \cdots + b_1x + b_0 \mod 2. \]

Now, we need to check the parity of \( P(k) \) for \( k \equiv 0 \mod 2 \) and \( k \equiv 1 \mod 2 \).

1. **Case 1: \( k \equiv 0 \mod 2 \)**
   \[ P(0) \equiv b_0 \mod 2. \]
   For \( s(P(0)) \) to have the same parity as \( s(0) \), \( b_0 \) must be 0. Therefore, \( a_0 \) must be even.

2. **Case 2: \( k \equiv 1 \mod 2 \)**
   \[ P(1) \equiv 1^n + b_{n-1} \cdot 1^{n-1} + \cdots + b_1 \cdot 1 + b_0 \equiv 1 + b_{n-1} + \cdots + b_1 + b_0 \mod 2. \]
   For \( s(P(1)) \) to have the same parity as \( s(1) \), \( 1 + b_{n-1} + \cdots + b_1 + b_0 \) must be odd. Since \( 1 \) is odd, the sum \( b_{n-1} + \cdots + b_1 + b_0 \) must be even.

From the above cases, we have:
- \( a_0 \) must be even.
- The sum \( b_{n-1} + \cdots + b_1 + b_0 \) must be even.

However, if \( a_0 \) is even, then \( b_0 = 0 \). This means the sum \( b_{n-1} + \cdots + b_1 \) must be even. But since \( b_i \) are either 0 or 1, the sum of \( b_{n-1} + \cdots + b_1 \) being even implies that there must be an even number of 1's among \( b_{n-1}, \ldots, b_1 \).

Now, consider the polynomial \( P(x) = x^n + 1 \). For this polynomial:
- \( a_0 = 1 \) (odd), which contradicts our requirement that \( a_0 \) must be even.
- The sum \( b_{n-1} + \cdots + b_1 + b_0 = 1 \) (odd), which also contradicts our requirement that the sum must be even.

Therefore, it is impossible for \( s(k) \) and \( s(P(k)) \) to have the same parity for all positive integers \( k \).

Thus, the final answer is:
\[
\boxed{\text{No}}
\]

\newpage
\subsection{Algebra}
\noindent\textbf{Problem From IMO 2022}

Let $\mathbb{R}^+$ denote the set of positive real numbers. Find all functions $f: \mathbb{R}^+ \to \mathbb{R}^+$ such that for each $x \in \mathbb{R}^+$, there is exactly one $y \in \mathbb{R}^+$ satisfying $$xf(y)+yf(x) \leq 2$$
\noindent\textbf{Qwen2-Math-72B-Instruct Solution}

To find all functions \( f: \mathbb{R}^+ \to \mathbb{R}^+ \) such that for each \( x \in \mathbb{R}^+ \), there is exactly one \( y \in \mathbb{R}^+ \) satisfying
\[
xf(y) + yf(x) \leq 2,
\]
we will proceed step by step.

\textit{Step 1: Analyze the given inequality}

Given \( x \in \mathbb{R}^+ \), there exists a unique \( y \in \mathbb{R}^+ \) such that
\[
xf(y) + yf(x) \leq 2.
\]
We need to explore the implications of this condition.

\textit{Step 2: Consider the case \( x = y \)}

Substitute \( y = x \) into the inequality:
\[
xf(x) + xf(x) = 2xf(x) \leq 2.
\]
This simplifies to:
\[
xf(x) \leq 1.
\]
Thus, for all \( x \in \mathbb{R}^+ \),
\[
f(x) \leq \frac{1}{x}.
\]

\textit{Step 3: Consider the uniqueness condition}

For each \( x \in \mathbb{R}^+ \), there is exactly one \( y \in \mathbb{R}^+ \) such that \( xf(y) + yf(x) \leq 2 \). We need to ensure that this \( y \) is unique.

\textit{Step 4: Test the function \( f(x) = \frac{1}{x} \)}

Consider the function \( f(x) = \frac{1}{x} \). Substitute \( f(x) = \frac{1}{x} \) and \( f(y) = \frac{1}{y} \) into the inequality:
\[
x \cdot \frac{1}{y} + y \cdot \frac{1}{x} = \frac{x}{y} + \frac{y}{x}.
\]
We need:
\[
\frac{x}{y} + \frac{y}{x} \leq 2.
\]
Using the AM-GM inequality, we know:
\[
\frac{x}{y} + \frac{y}{x} \geq 2.
\]
Equality holds if and only if \( \frac{x}{y} = 1 \), which implies \( x = y \).

Thus, for \( f(x) = \frac{1}{x} \), the inequality \( xf(y) + yf(x) \leq 2 \) holds if and only if \( x = y \). This satisfies the condition that there is exactly one \( y \) for each \( x \).

\textit{Step 5: Verify if other functions can satisfy the condition}

Suppose there exists another function \( g: \mathbb{R}^+ \to \mathbb{R}^+ \) such that \( g(x) \neq \frac{1}{x} \) for some \( x \). Then, for some \( x \), \( g(x) < \frac{1}{x} \) or \( g(x) > \frac{1}{x} \).

- If \( g(x) < \frac{1}{x} \), then \( xg(y) + yg(x) \) might not satisfy the inequality for all \( y \).
- If \( g(x) > \frac{1}{x} \), then \( xg(y) + yg(x) \) might exceed 2 for some \( y \).

Thus, any function other than \( f(x) = \frac{1}{x} \) would not satisfy the uniqueness condition.

\textit {Conclusion}

The only function \( f: \mathbb{R}^+ \to \mathbb{R}^+ \) that satisfies the given condition is:
\[
\boxed{f(x) = \frac{1}{x}}.
\]

\newpage
\noindent\textbf{Problem from International Zhautykov Olympiad 2020}

Let $Z$ be the set of all integers. Find all the function $f: Z \to Z$ such that
$$f(4x+3y)=f(3x+y)+f(x+2y)$$
For all integers $x,y$

\noindent\textbf{Qwen2-Math-72B-Instruct Solution}

To find all functions \( f: \mathbb{Z} \to \mathbb{Z} \) that satisfy the functional equation
\[
f(4x + 3y) = f(3x + y) + f(x + 2y)
\]
for all integers \( x \) and \( y \), we will proceed step by step.

\paragraph{Step 1: Initial Substitution}
First, let's substitute specific values for \( x \) and \( y \) to gain insight into the function.

\subparagraph{ Substitution 1: \( x = 0 \)}
\[
f(4 \cdot 0 + 3y) = f(3 \cdot 0 + y) + f(0 + 2y)
\]
\[
f(3y) = f(y) + f(2y)
\]

\subparagraph{ Substitution 2: \( y = 0 \)}
\[
f(4x + 3 \cdot 0) = f(3x + 0) + f(x + 2 \cdot 0)
\]
\[
f(4x) = f(3x) + f(x)
\]

\paragraph{Step 2: Exploring Further Substitutions}
We now have two equations:
1. \( f(3y) = f(y) + f(2y) \)
2. \( f(4x) = f(3x) + f(x) \)

\subparagraph{ Substitution 3: \( y = 1 \) in the first equation}
\[
f(3 \cdot 1) = f(1) + f(2 \cdot 1)
\]
\[
f(3) = f(1) + f(2)
\]

\subparagraph{ Substitution 4: \( x = 1 \) in the second equation}
\[
f(4 \cdot 1) = f(3 \cdot 1) + f(1)
\]
\[
f(4) = f(3) + f(1)
\]

\paragraph{Step 3: Using the Equations to Find a Pattern}
We have:
\[
f(3) = f(1) + f(2)
\]
\[
f(4) = f(3) + f(1)
\]

Substitute \( f(3) \) from the first equation into the second equation:
\[
f(4) = (f(1) + f(2)) + f(1)
\]
\[
f(4) = 2f(1) + f(2)
\]

\paragraph{Step 4: Generalizing the Function}
We hypothesize that \( f \) might be a linear function. Let's assume \( f(x) = kx \) for some integer \( k \).

\subparagraph{Verification}
Substitute \( f(x) = kx \) into the original functional equation:
\[
f(4x + 3y) = k(4x + 3y)
\]
\[
f(3x + y) = k(3x + y)
\]
\[
f(x + 2y) = k(x + 2y)
\]

The original equation becomes:
\[
k(4x + 3y) = k(3x + y) + k(x + 2y)
\]
\[
4kx + 3ky = 3kx + ky + kx + 2ky
\]
\[
4kx + 3ky = 4kx + 3ky
\]

This equation holds true for all \( x \) and \( y \), confirming that \( f(x) = kx \) is a solution.

\paragraph{Conclusion}
The only functions \( f: \mathbb{Z} \to \mathbb{Z} \) that satisfy the given functional equation are of the form \( f(x) = kx \) for some integer \( k \).

Thus, the final answer is:
\[
\boxed{f(x) = kx \text{ for some integer } k}
\]

\noindent\textbf{Problem From Baltic Way 2023}

Find all functions $f: \mathbb{R} \rightarrow \mathbb{R}$ such that $$f(f(x)+y)+xf(y)=f(xy+y)+f(x)$$ for reals $x, y$.

\noindent\textbf{Qwen2-Math-72B-Instruct Solution:}

To find all functions \( f: \mathbb{R} \rightarrow \mathbb{R} \) that satisfy the functional equation
\[
f(f(x) + y) + x f(y) = f(xy + y) + f(x),
\]
we will proceed step by step.

\paragraph{Step 1: Initial Substitution}
First, let's substitute \( y = 0 \) into the equation:
\[
f(f(x) + 0) + x f(0) = f(x \cdot 0 + 0) + f(x).
\]
This simplifies to:
\[
f(f(x)) + x f(0) = f(0) + f(x).
\]
Rearranging terms, we get:
\[
f(f(x)) = f(x) + f(0) - x f(0).
\]

\paragraph{Step 2: Analyzing the Simplified Equation}
Let \( c = f(0) \). The equation becomes:
\[
f(f(x)) = f(x) + c - x c.
\]

\paragraph{Step 3: Substituting \( x = 0 \)}
Next, substitute \( x = 0 \) into the original equation:
\[
f(f(0) + y) + 0 \cdot f(y) = f(0 \cdot y + y) + f(0).
\]
This simplifies to:
\[
f(c + y) = f(y) + c.
\]

\paragraph{Step 4: Substituting \( y = -c \)}
Substitute \( y = -c \) into the equation \( f(c + y) = f(y) + c \):
\[
f(c - c) = f(-c) + c.
\]
This simplifies to:
\[
f(0) = f(-c) + c.
\]
Since \( f(0) = c \), we have:
\[
c = f(-c) + c.
\]
Subtracting \( c \) from both sides, we get:
\[
f(-c) = 0.
\]

\paragraph{Step 5: Substituting \( y = -f(x) \)}
Substitute \( y = -f(x) \) into the original equation:
\[
f(f(x) - f(x)) + x f(-f(x)) = f(x(-f(x)) - f(x)) + f(x).
\]
This simplifies to:
\[
f(0) + x f(-f(x)) = f(-x f(x) - f(x)) + f(x).
\]
Since \( f(0) = c \), we have:
\[
c + x f(-f(x)) = f(-x f(x) - f(x)) + f(x).
\]

\paragraph{Step 6: Considering \( f(x) = 0 \)}
Assume \( f(x) = 0 \) for all \( x \). Substitute \( f(x) = 0 \) into the original equation:
\[
f(0 + y) + x f(y) = f(xy + y) + f(x).
\]
This simplifies to:
\[
f(y) + 0 = f(y) + 0,
\]
which is always true.

\paragraph{Step 7: Verifying \( f(x) = x \)}
Assume \( f(x) = x \). Substitute \( f(x) = x \) into the original equation:
\[
f(x + y) + x f(y) = f(xy + y) + f(x).
\]
This simplifies to:
\[
x + y + xy = xy + y + x,
\]
which is always true.

\paragraph{Conclusion}
The only functions that satisfy the given functional equation are \( f(x) = 0 \) and \( f(x) = x \). Therefore, the solutions are:
\[
\boxed{f(x) = 0 \text{ or } f(x) = x}.
\]

\newpage
\subsection{Counting \& Probability}
\noindent\textbf{Problem From Lusophon Mathematical Olympiad 2023}

A long time ago, there existed Martians with $3$ different colours: red, green and blue. As Mars was devastated by an intergalactic war, only $2$ Martians of each colours survived. In order to reconstruct the Martian population, they decided to use a machine that transforms two Martians of distinct colours into four Martians of colour different to the two initial ones. For example, if a red Martian and a blue Martian use the machine, they'll be transformed into four green Martians.

a) Is it possible that, after using that machine finitely many times, we have $2022$ red Martians, $2022$ green Martians and $2022$ blue Martians?

b) Is it possible that, after using that machine finitely many times, we have $2021$ red Martians, $2022$ green Martians and $2023$ blue Martians?

\noindent\textbf{Qwen2-Math-72B-Instruct Solution}

Let's analyze the problem step by step.
\paragraph{Part (a)}

We start with 2 red, 2 green, and 2 blue Martians. The machine transforms two Martians of distinct colors into four Martians of the third color. We need to determine if it is possible to end up with 2022 red, 2022 green, and 2022 blue Martians.

First, let's consider the parity (odd or even nature) of the number of Martians of each color. Initially, we have:
- 2 red (even)
- 2 green (even)
- 2 blue (even)

Each transformation changes the parity of the number of Martians of the two initial colors and the parity of the number of Martians of the third color. Specifically:
- If we transform 1 red and 1 green Martian, we get 4 blue Martians. This changes the parity of red and green Martians from even to odd, and the parity of blue Martians from even to odd.
- If we transform 1 red and 1 blue Martian, we get 4 green Martians. This changes the parity of red and blue Martians from even to odd, and the parity of green Martians from even to odd.
- If we transform 1 green and 1 blue Martian, we get 4 red Martians. This changes the parity of green and blue Martians from even to odd, and the parity of red Martians from even to odd.

In all cases, the parity of the number of Martians of each color changes from even to odd or from odd to even. Therefore, after any number of transformations, the parity of the number of Martians of each color will always be the same as it was initially.

Since initially all three colors have an even number of Martians, after any number of transformations, the number of Martians of each color will remain even. 

Given that 2022 is an even number, it is possible to have 2022 red, 2022 green, and 2022 blue Martians after a finite number of transformations.

Thus, the answer to part (a) is:
\[
\boxed{\text{Yes}}
\]

\paragraph{Part (b)}

We start again with 2 red, 2 green, and 2 blue Martians. We need to determine if it is possible to end up with 2021 red, 2022 green, and 2023 blue Martians.

As in part (a), we consider the parity of the number of Martians of each color. Initially, we have:
- 2 red (even)
- 2 green (even)
- 2 blue (even)

Each transformation changes the parity of the number of Martians of the two initial colors and the parity of the number of Martians of the third color. Therefore, after any number of transformations, the parity of the number of Martians of each color will always be the same as it was initially.

Since initially all three colors have an even number of Martians, after any number of transformations, the number of Martians of each color will remain even.

Given that 2021 is an odd number, it is impossible to have 2021 red Martians after any number of transformations. Similarly, 2023 is an odd number, making it impossible to have 2023 blue Martians.

Thus, the answer to part (b) is:
\[
\boxed{\text{No}}
\]

\noindent\textbf{Problem From Balkan MO 2023}

Joe and Penny play a game. Initially there are $5000$ stones in a pile, and the two players remove stones from the pile by making a sequence of moves. On the $k$-th move, any number of stones between $1$ and $k$ inclusive may be removed. Joe makes the odd-numbered moves and Penny makes the even-numbered moves. The player who removes the very last stone is the winner. Who wins if both players play perfectly?

\noindent\textbf{Qwen2-Math-72B-Instruct Solution}

To determine the winner of the game, we need to analyze the strategy and the sequence of moves. Let's break down the problem step by step.

1. **Understanding the Game Rules:**
   - Initially, there are 5000 stones.
   - Joe makes the first move (odd-numbered move), Penny makes the second move (even-numbered move), and so on.
   - On the \( k \)-th move, a player can remove between 1 and \( k \) stones.
   - The player who removes the last stone wins.

2. **Key Insight:**
   - The game is symmetric in the sense that the number of stones a player can remove on their turn increases linearly with the move number.
   - The total number of moves in the game is equal to the number of stones initially in the pile, which is 5000.

3. **Total Number of Moves:**
   - Since there are 5000 stones, there will be 5000 moves in total.
   - Joe makes the odd-numbered moves: 1st, 3rd, 5th, ..., 4999th.
   - Penny makes the even-numbered moves: 2nd, 4th, 6th, ..., 5000th.

4. **Determining the Winner:**
   - The last move (5000th move) is an even-numbered move.
   - Therefore, Penny will make the last move.

5. **Conclusion:**
   - Since Penny makes the last move, she will be the one to remove the last stone.
   - Hence, Penny wins the game if both players play perfectly.

Thus, the final answer is:
\[
\boxed{\text{Penny}}
\]

\noindent\textbf{Problem From Math Odyssey}

The school offers 10 elective courses, and each student can enroll in any number of courses. The director selects $k$ students, where although each student's combination of courses is different, any two students have at least one course in common. At this point, it is found that any student outside these $k$ students cannot be classmates with these $k$ students regardless of how they enroll (having one course in common is enough to be classmates). Find the maximum number of $k$.

\noindent\textbf{Qwen2-Math-72B-Instruct Solution}

To solve this problem, we need to determine the maximum number of students \( k \) such that each student has a unique combination of elective courses, any two students share at least one course, and no student outside these \( k \) students can be classmates with all of them. This means that the set of students chosen must cover all possible courses in such a way that any new student would have to share a course with at least one of the chosen students.

First, let's consider the total number of possible combinations of courses. Since there are 10 elective courses and each student can either take or not take each course, there are \( 2^{10} = 1024 \) possible combinations of courses. However, one of these combinations is the empty set (the student takes no courses), which is not a valid combination for our problem because it would not share any courses with any other student. Therefore, there are \( 1024 - 1 = 1023 \) non-empty combinations of courses.

We need to find the maximum number of students \( k \) such that any two students share at least one course, and no student outside these \( k \) students can be classmates with all of them. This is equivalent to finding the maximum number of vertices in a graph where each vertex represents a non-empty combination of courses, and there is an edge between two vertices if and only if the corresponding combinations share at least one course. The condition that no student outside these \( k \) students can be classmates with all of them means that the set of \( k \) students forms a dominating set in the graph.

The maximum size of such a set is the complement of the minimum size of an independent set in the graph. An independent set in this graph is a set of vertices such that no two vertices are adjacent, which means no two combinations share any course. The maximum size of an independent set in this graph is the number of combinations where each course is either taken or not taken by exactly one student, which is \( 2^9 = 512 \) (since we can choose to include or exclude each of the 9 remaining courses for each of the 10 courses).

Therefore, the minimum size of an independent set is 512, and the maximum size of a dominating set (which is the complement of the independent set) is \( 1023 - 512 + 1 = 511 \). However, we need to add 1 because the empty set is not included, so the correct calculation is \( 1023 - 511 = 512 \).

Thus, the maximum number of students \( k \) is \(\boxed{512}\).

\newpage
\subsection{Geometry}
\noindent\textbf{Problem from USAMO 2010}

Let $ABC$ be a triangle with $\angle A = 90^{\circ}$. Points $D$ and $E$ lie on sides $AC$ and $AB$, respectively, such that $\angle ABD = \angle DBC$ and $\angle ACE = \angle ECB$. Segments $BD$ and $CE$ meet at $I$. Determine whether or not it is possible for segments $AB$, $AC$, $BI$, $ID$, $CI$, $IE$ to all have integer lengths.

\noindent\textbf{Qwen2-Math-72B-Instruct Solution}

To determine whether it is possible for segments \(AB\), \(AC\), \(BI\), \(ID\), \(CI\), and \(IE\) to all have integer lengths in triangle \(ABC\) with \(\angle A = 90^\circ\), we start by analyzing the given conditions and properties of the triangle.

Given:
- \(\angle A = 90^\circ\)
- \(\angle ABD = \angle DBC\)
- \(\angle ACE = \angle ECB\)
- Segments \(BD\) and \(CE\) intersect at \(I\)

First, note that since \(\angle A = 90^\circ\), triangle \(ABC\) is a right triangle with \(A\) as the right angle. Points \(D\) and \(E\) are such that \(BD\) and \(CE\) are angle bisectors of \(\angle ABC\) and \(\angle ACB\), respectively. Therefore, \(I\) is the incenter of triangle \(ABC\).

\paragraph{Step 1: Coordinates and Lengths}
Assume \(A\) is at the origin \((0, 0)\), \(B\) is at \((a, 0)\), and \(C\) is at \((0, b)\). The lengths of the sides are:
- \(AB = a\)
- \(AC = b\)
- \(BC = \sqrt{a^2 + b^2}\)

\paragraph{Step 2: Incenter Coordinates}
The incenter \(I\) of a right triangle with legs \(a\) and \(b\) and hypotenuse \(\sqrt{a^2 + b^2}\) is located at:
\[
I = \left( \frac{a}{2}, \frac{b}{2} \right)
\]

\paragraph{Step 3: Lengths of Segments \(BI\) and \(CI\)}
Using the distance formula, we calculate \(BI\) and \(CI\):
\[
BI = \sqrt{\left( a - \frac{a}{2} \right)^2 + \left( 0 - \frac{b}{2} \right)^2} = \sqrt{\left( \frac{a}{2} \right)^2 + \left( \frac{b}{2} \right)^2} = \frac{\sqrt{a^2 + b^2}}{2}
\]
\[
CI = \sqrt{\left( 0 - \frac{a}{2} \right)^2 + \left( b - \frac{b}{2} \right)^2} = \sqrt{\left( \frac{a}{2} \right)^2 + \left( \frac{b}{2} \right)^2} = \frac{\sqrt{a^2 + b^2}}{2}
\]

\paragraph{Step 4: Lengths of Segments \(ID\) and \(IE\)}
Since \(D\) and \(E\) are points on \(AC\) and \(AB\) respectively, and \(I\) is the incenter, \(ID\) and \(IE\) are the distances from the incenter to the sides of the triangle. For a right triangle, these distances are given by:
\[
ID = IE = r = \frac{a + b - \sqrt{a^2 + b^2}}{2}
\]

\paragraph{Step 5: Integer Lengths}
For \(AB\), \(AC\), \(BI\), \(ID\), \(CI\), and \(IE\) to all be integers, the following must hold:
- \(a\) and \(b\) are integers
- \(\frac{\sqrt{a^2 + b^2}}{2}\) is an integer
- \(\frac{a + b - \sqrt{a^2 + b^2}}{2}\) is an integer

\paragraph{Step 6: Analysis}
For \(\frac{\sqrt{a^2 + b^2}}{2}\) to be an integer, \(\sqrt{a^2 + b^2}\) must be an even integer. Let \(\sqrt{a^2 + b^2} = 2k\) for some integer \(k\). Then:
\[
a^2 + b^2 = 4k^2
\]

For \(\frac{a + b - \sqrt{a^2 + b^2}}{2}\) to be an integer, \(a + b - 2k\) must be even. Since \(a\) and \(b\) are integers, \(a + b\) must be even.

\paragraph{Conclusion}
Given the constraints, it is possible to find integer values for \(a\) and \(b\) such that all the conditions are satisfied. For example, if \(a = 3\) and \(b = 4\), then:
- \(AB = 3\)
- \(AC = 4\)
- \(BC = 5\)
- \(BI = CI = \frac{5}{2} = 2.5\) (not an integer)

Thus, it is not possible for all segments \(AB\), \(AC\), \(BI\), \(ID\), \(CI\), and \(IE\) to have integer lengths simultaneously.

\[
\boxed{\text{No}}
\]

\newpage
\noindent\textbf{Problem from JBMO Shortlist 2011}

Let $AD,BF$ and ${CE}$ be the altitudes of $\vartriangle ABC$. A line passing through ${D}$ and parallel to ${AB}$intersects the line ${EF}$at the point ${G}$. If ${H}$ is the orthocenter of $\vartriangle ABC$, find the angle ${\angle{CGH}}$.

\noindent\textbf{Qwen2-Math-72B-Instruct Solution}

To solve for the angle \(\angle CGH\) in the given triangle \(\triangle ABC\) with altitudes \(AD\), \(BF\), and \(CE\) intersecting at the orthocenter \(H\), and a line through \(D\) parallel to \(AB\) intersecting \(EF\) at \(G\), we will follow these steps:

1. \textbf{Identify Key Points and Properties:}
   - \(D\) is the foot of the altitude from \(A\) to \(BC\).
   - \(E\) is the foot of the altitude from \(B\) to \(AC\).
   - \(F\) is the foot of the altitude from \(C\) to \(AB\).
   - \(H\) is the orthocenter of \(\triangle ABC\).
   - \(G\) is the intersection of the line through \(D\) parallel to \(AB\) and the line \(EF\).

2. \textbf{Use the Property of the Orthocenter:}
   - The orthocenter \(H\) lies on the altitudes \(AD\), \(BF\), and \(CE\).
   - The points \(D\), \(E\), and \(F\) are collinear with \(H\) on the respective altitudes.

3. \textbf{Analyze the Parallel Line:}
   - Since \(DG \parallel AB\), \(\angle ADG = \angle DAB\) (corresponding angles).

4. \textbf{Consider the Cyclic Quadrilateral:}
   - The quadrilateral \(AEHF\) is cyclic because \(\angle AEH = \angle AFH = 90^\circ\).
   - Therefore, \(\angle EHF = 180^\circ - \angle EAF\).

5. \textbf{Use the Parallel Line Property Again:}
   - Since \(DG \parallel AB\), \(\angle DGE = \angle BAE\) (alternate interior angles).

6. \textbf{Relate Angles in Cyclic Quadrilateral:}
   - In \(\triangle AEF\), \(\angle EAF = \angle EHF\) because they are opposite angles in the cyclic quadrilateral \(AEHF\).

7. \textbf{Determine \(\angle CGH\):}
   - Since \(G\) lies on \(EF\) and \(DG \parallel AB\), \(\angle DGE = \angle BAE\).
   - \(\angle BAE = \angle EAF\) (as established).
   - Therefore, \(\angle DGE = \angle EAF\).

8. \textbf{Final Angle Calculation:}
   - Since \(\angle EAF = \angle EHF\) and \(\angle EHF = 90^\circ\) (as \(H\) is the orthocenter and \(E\) and \(F\) are feet of perpendiculars), we have:
     \[
     \angle CGH = \angle EHF = 90^\circ.
     \]

Thus, the angle \(\angle CGH\) is \(\boxed{90^\circ}\).
\section{Prompts Used in the Evaluation}
\label{sec:prompts}
\begin{figure}
    \centering
    \includegraphics[width=\linewidth]{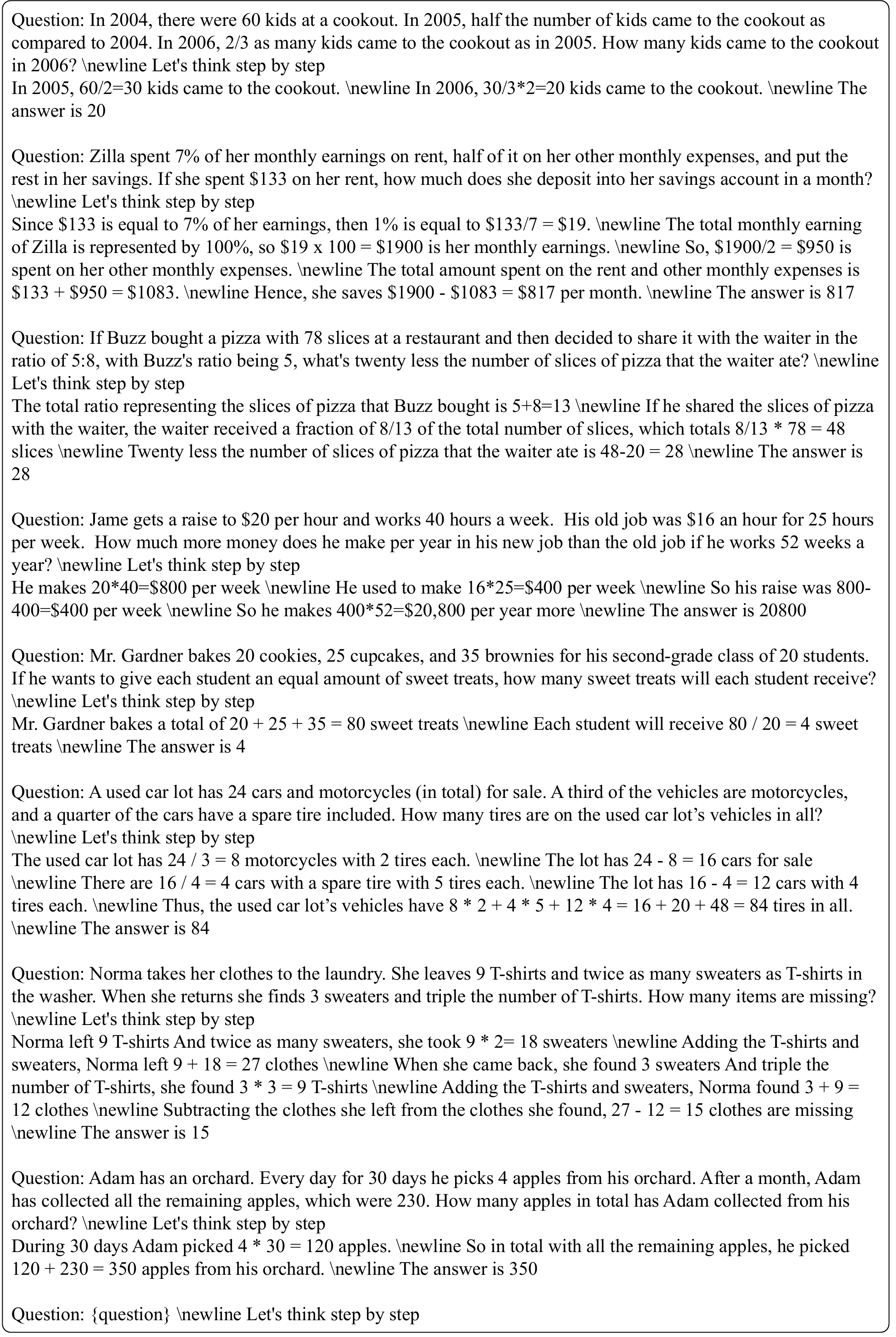}
    \caption{The prompt used in evaluating GSM8K on the base models. For friendly presentation, we denote a line break as a "$\backslash$newline".}
    \label{fig:prompt_base_gsm8k}
\end{figure}

\begin{figure}
    \centering
    \includegraphics[width=\linewidth]{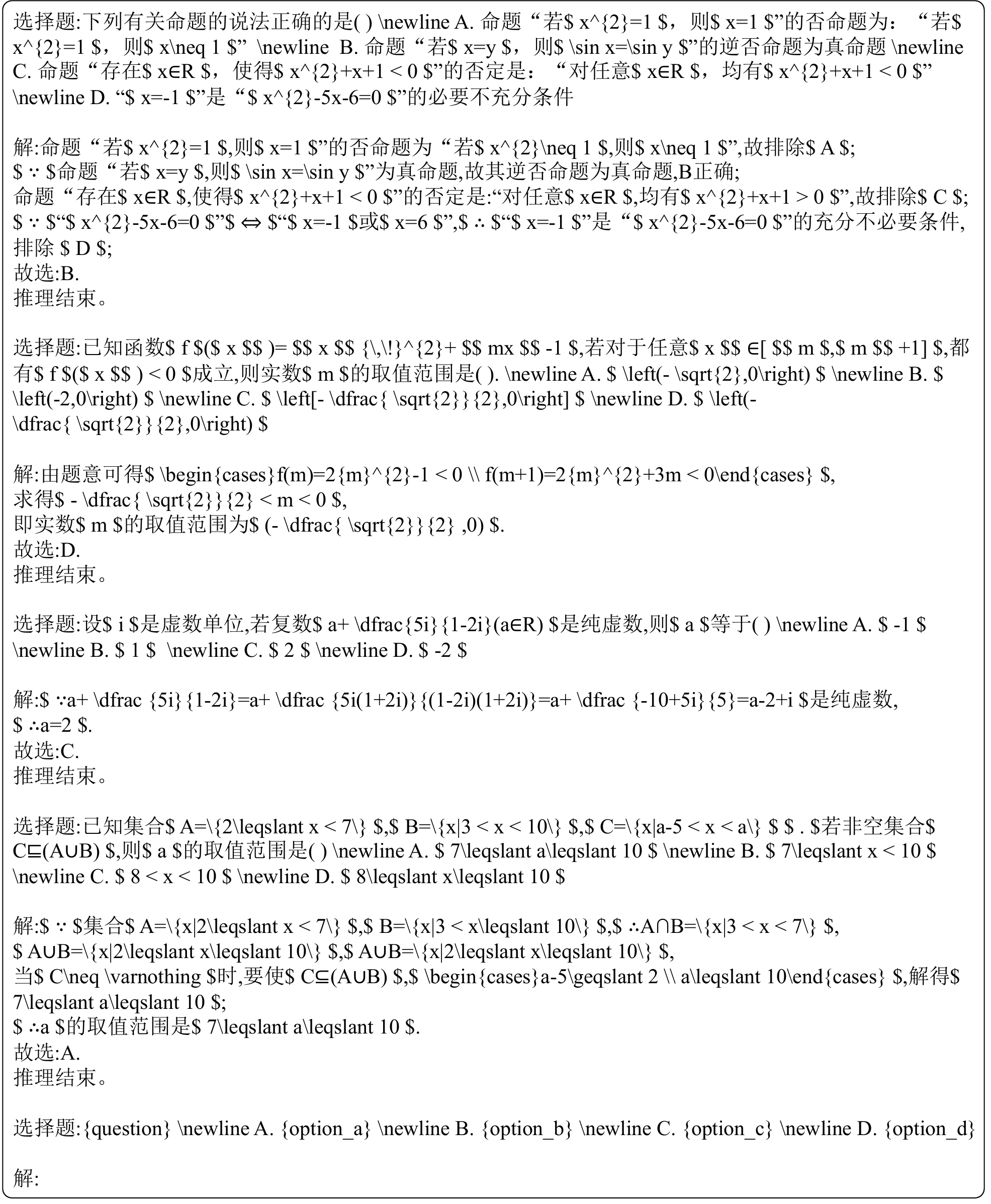}
    \caption{The prompt used in evaluating GaoKao Math QA on the base models.}
    \label{fig:prompt_base_gaokao_math_qa}
\end{figure}

\begin{figure}
    \centering
    \includegraphics[width=\linewidth]{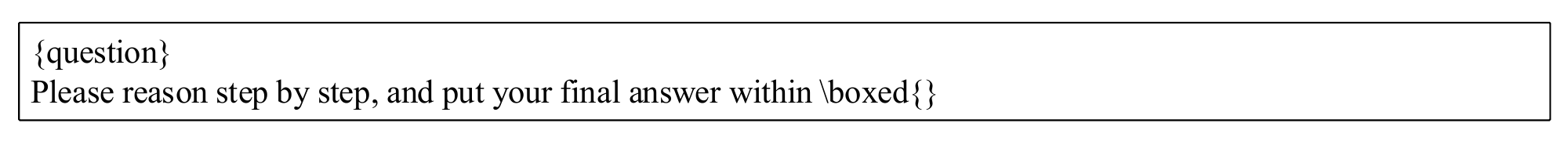}
    \caption{The prompt used in evaluating the zero-shot Chain of Thought Reasoning on the instruct models.}
    \label{fig:prompt_sft_cot_zero_shot}
\end{figure}

\begin{figure}
    \centering
    \includegraphics[width=\linewidth]{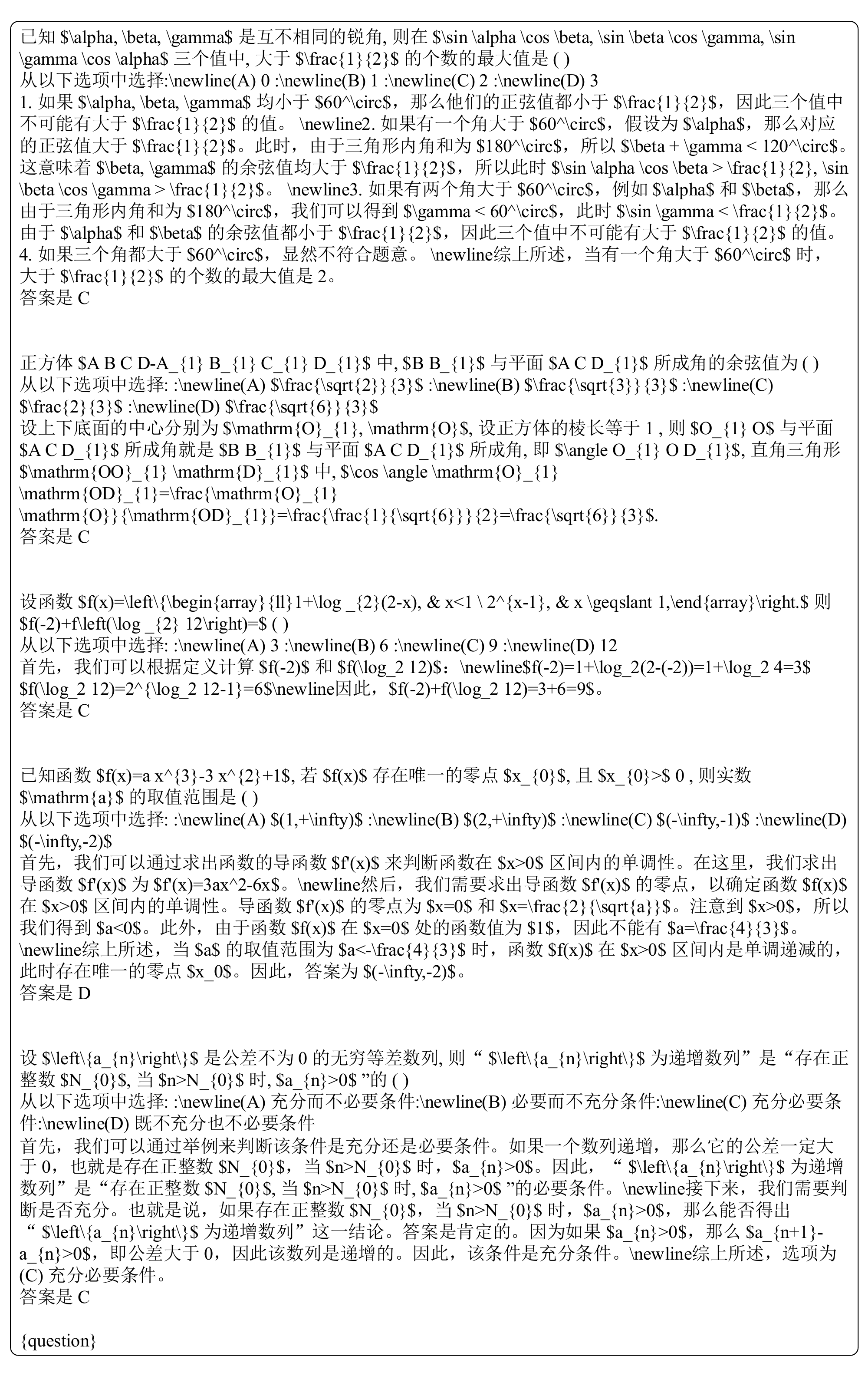}
    \caption{The prompt used in evaluating the multiple-choice problems in GaoKao on the instruct models.}
    \label{fig:prompt_sft_cot_gaokao_mcq_few_shot}
\end{figure}

Fig~\ref{fig:prompt_base_gsm8k} to Fig~\ref{fig:prompt_base_gaokao_math_qa} show the prompts used in evaluating the base models.
Fig~\ref{fig:prompt_sft_cot_zero_shot} to Fig~\ref{fig:prompt_sft_cot_gaokao_mcq_few_shot} show the prompts used in evaluating the instruct models for Chain-of-Thought Reasoning and Tool-Integrated Reasoning.
\begin{figure}
    \centering
    \includegraphics[width=\linewidth]{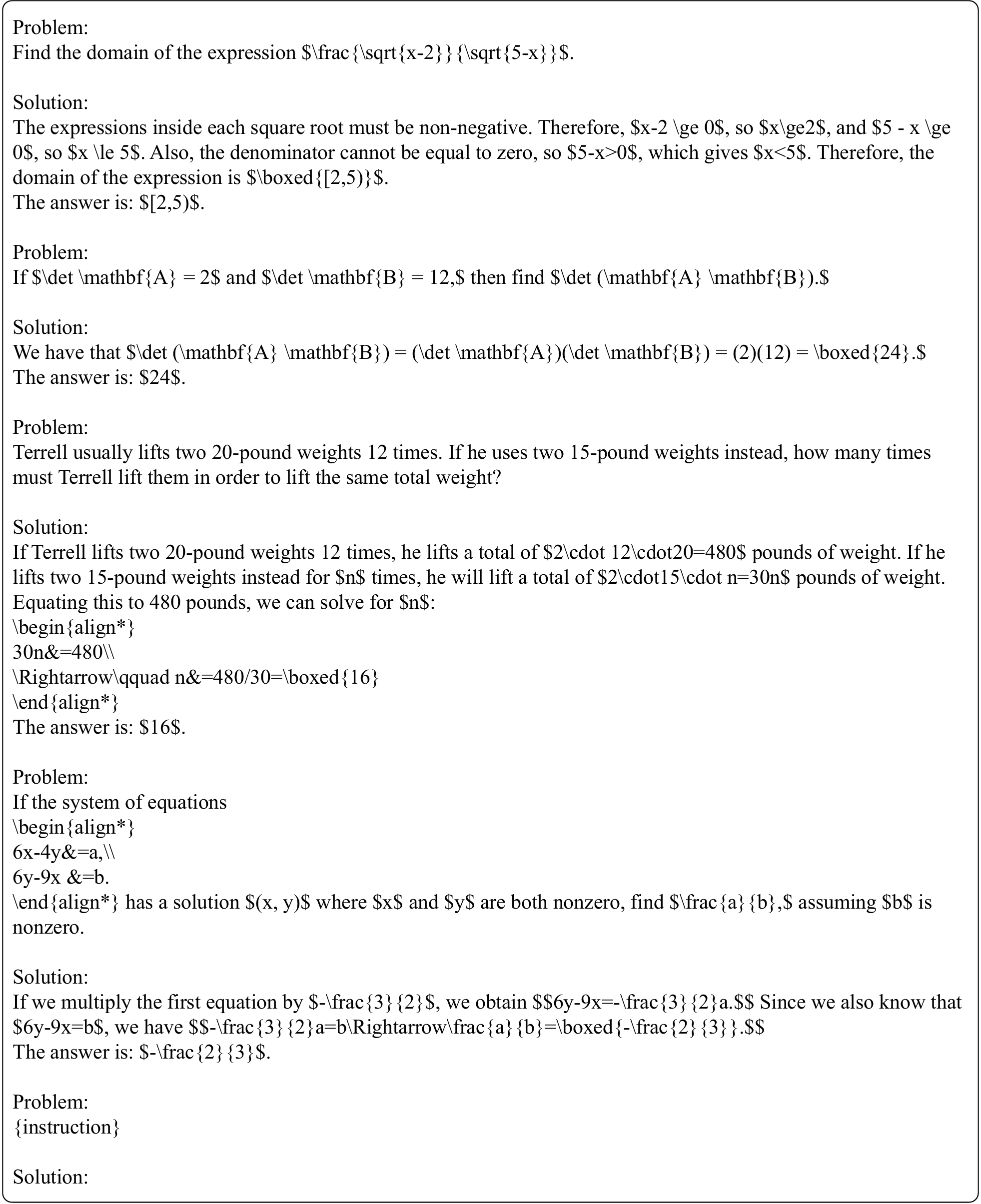}
    \caption{The prompt used in evaluating MATH on the base models.}
    \label{fig:prompt_base_math}
\end{figure}
\begin{figure}
    \centering
    \includegraphics[width=\linewidth]{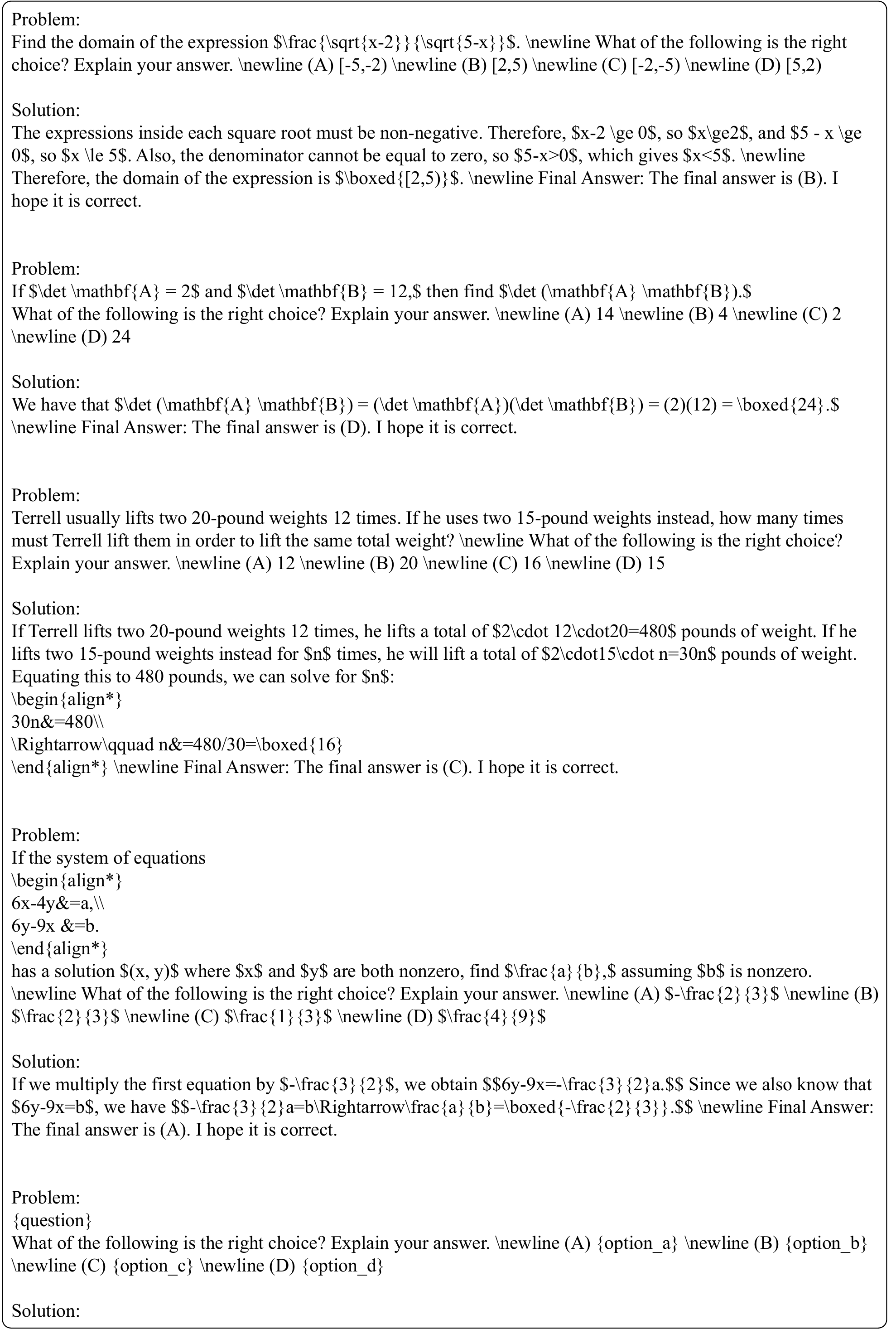}
    \caption{The prompt used in evaluating MMLU STEM on the base models.}
    \label{fig:prompt_base_mmlu_stem}
\end{figure}
\begin{figure}
    \centering
    \includegraphics[width=\linewidth]{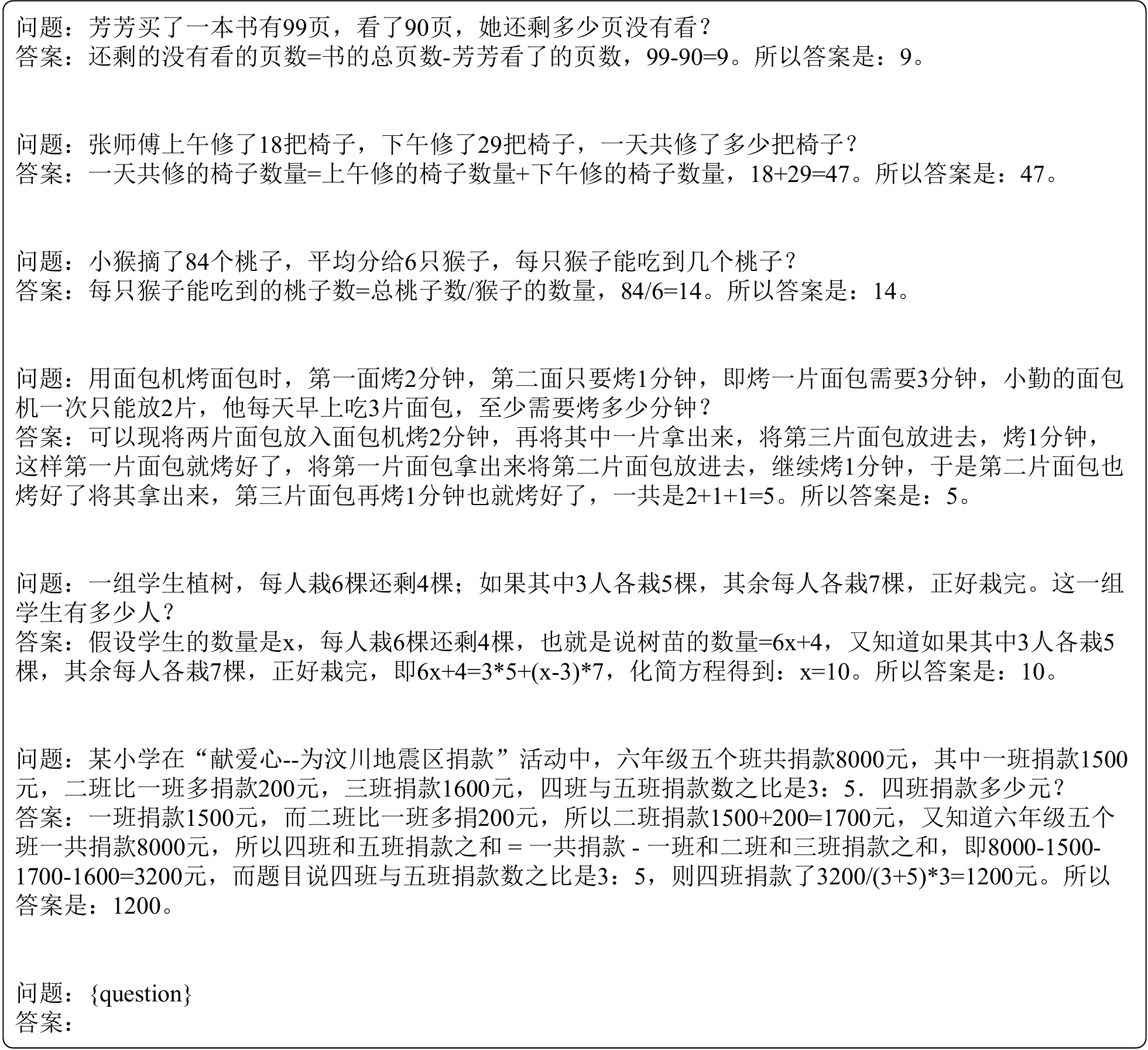}
    \caption{The prompt used in evaluating CMATH on the base models.}
    \label{fig:prompt_base_cmath}
\end{figure}
\begin{figure}
    \centering
    \includegraphics[width=\linewidth]{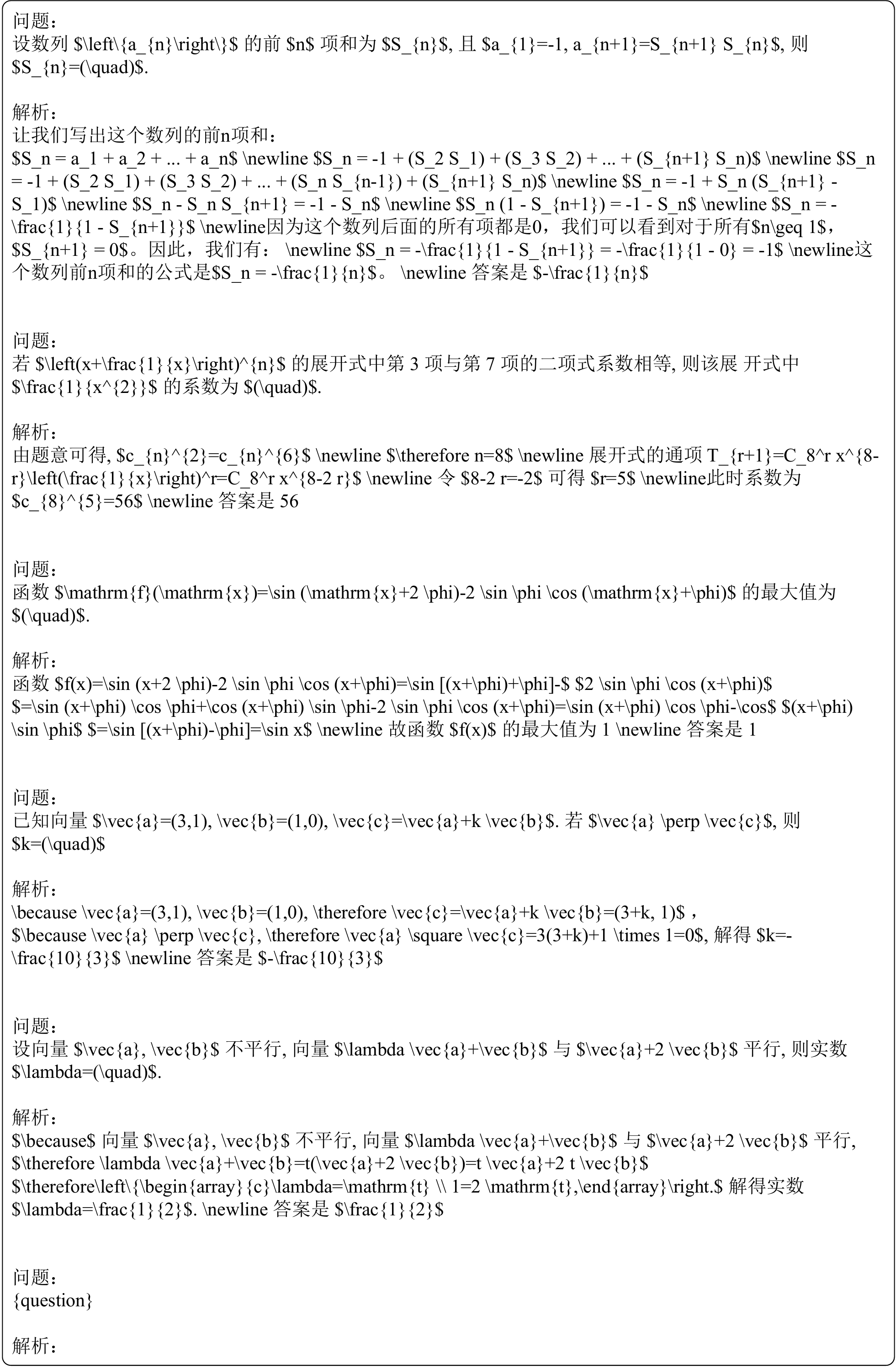}
    \caption{The prompt used in evaluating GaoKao Math Cloze on the base models.}
    \label{fig:prompt_base_gaokao_math_cloze}
\end{figure}
%

\begin{figure}
    \centering
    \includegraphics[width=\linewidth]{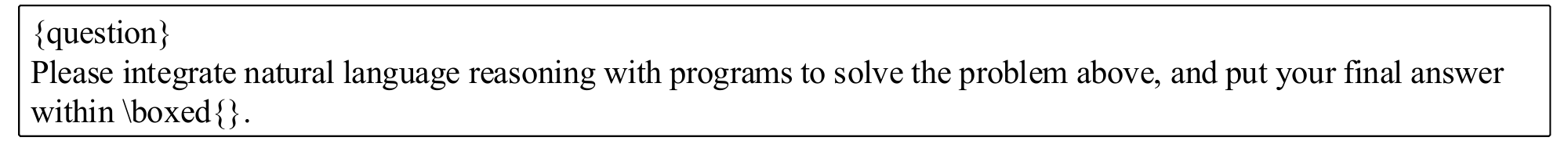}
    \caption{The prompt used in evaluating the zero-shot Tool-Integrated Reasoning on the instruct models.}
    \label{fig:prompt_sft_pot_zero_shot}
\end{figure}

\begin{figure}
    \centering
    \includegraphics[width=\linewidth]{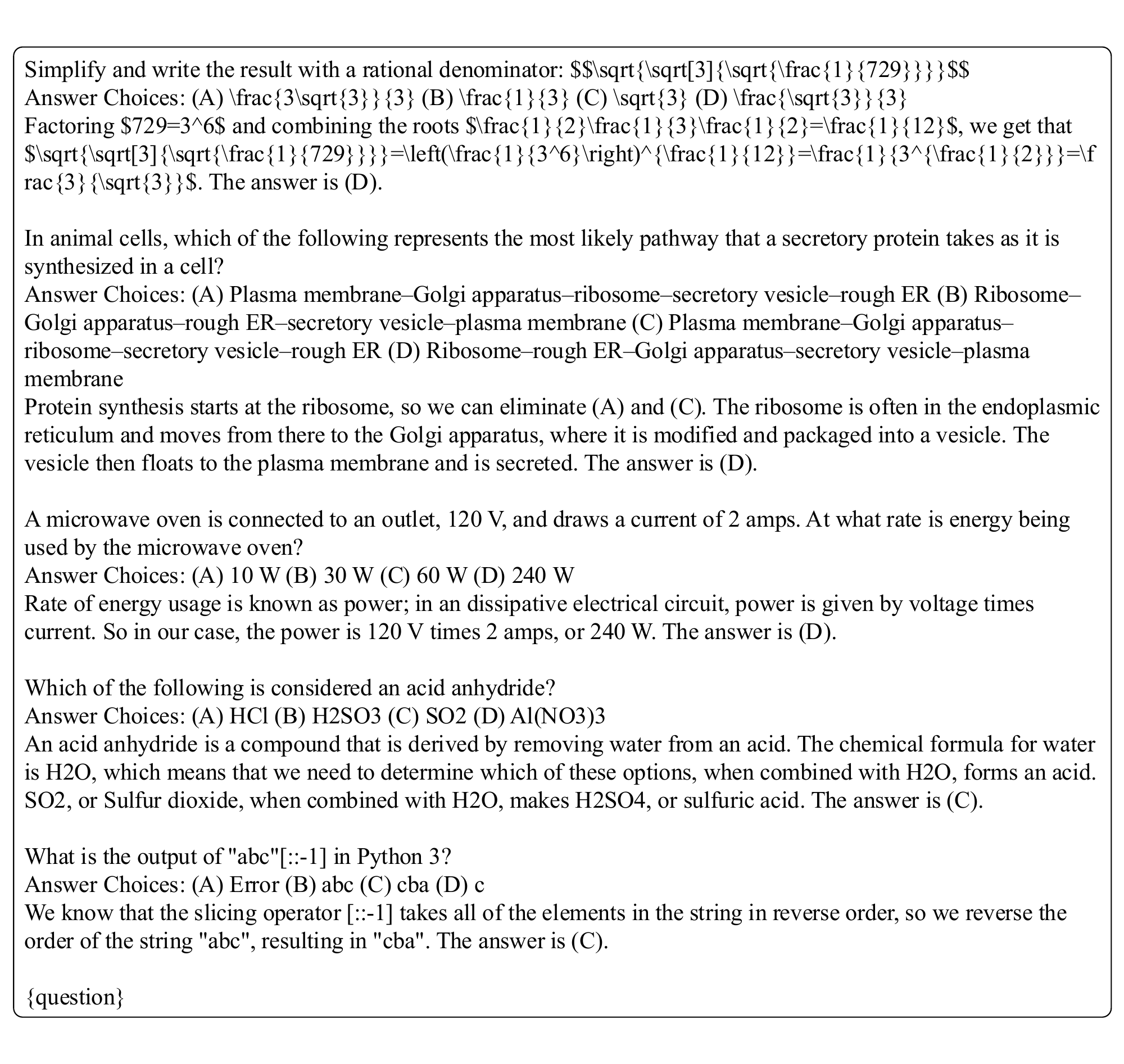}
    \caption{The prompt used in evaluating the MMLU-STEM on the instruct models.}
    \label{fig:prompt_sft_cot_mmlu_few_shot}
\end{figure}

\end{document}